\documentclass[twocolumn,preprint]{elsarticle}

\usepackage{amsmath}
\usepackage{epstopdf}
\usepackage{amssymb}
\usepackage{url}
\usepackage[normalem]{ulem}
\usepackage[usenames, dvipsnames]{color}
\usepackage{graphicx}
\usepackage{soul} 

\usepackage{color}
\usepackage{setspace}
\usepackage{rotating} 
\usepackage{rotating}
\usepackage{pdflscape}
\usepackage{tikz}
\usetikzlibrary{arrows}

\usepackage[english]{babel}
\usepackage[tworuled,linesnumbered]{algorithm2e}
\usepackage{blindtext}

\usepackage[ddmmyyyy,hhmmss]{datetime}

\newcommand{\jname}{Computational Biology and Chemistry (Elsevier), 2016 | \url{DOI:10.1016/j.compbiolchem.2016.01.008}}

\journal{\emph{\bf\jname}}

\newcommand{\ignore}[1]{}

\newcommand{\pct}{{\scriptsize \%}}
\newcommand{\fcc}{face-centered-cubic}

\newcommand{\hcc}{hydrophobic-core-center}

\newcommand{\aas}{amino acids}

\newcommand{\abi}{\emph{ab initio}}
\newcommand{\Abi}{\emph{Ab initio}}
\newcommand{\sota}{state-of-the-art}
\newcommand{\etal}{\emph{et al.}}
\newcommand{\pcode}{\textit{pseudocode}}

\newcommand{\fig}{Figure~}
\newcommand{\tab}{Table~}
\newcommand{\alg}{Algorithm~}

\newcommand{\lin}{Line~}

\newcommand{\ttx}{$20\times20$}

\newcommand{\ba}{\it 4RXN}
\newcommand{\bb}{\it 1ENH}
\newcommand{\bc}{\it 4PTI}
\newcommand{\bd}{\it 2IGD}
\newcommand{\be}{\it 1YPA}
\newcommand{\bfx}{\it 1R69}
\newcommand{\bg}{\it 1CTF}

\newcommand{\ca}{\it 3MX7}
\newcommand{\cb}{\it 3NBM}
\newcommand{\cc}{\it CMQO}
\newcommand{\cd}{\it 3MRO}
\newcommand{\ce}{\it 3PNX}

\newcommand{\da}{\it 2J61}
\newcommand{\db}{\it 2HFQ}

\newcommand{\rashid}{Mahmood A. Rashid}
\newcommand{\firas}{Firas Khatib}
\newcommand{\tamjid}{Md Tamjidul Hoque}
\newcommand{\abdul}{Abdul Sattar}
\newcommand{\sumaiya}{Sumaiya Iqbal}

\newcommand{\rashidE}{mahmood.rashid@usp.ac.fj}
\newcommand{\firasE}{fkhatib@umassd.edu}
\newcommand{\tamjidE}{thoque@uno.edu}
\newcommand{\abdulE}{a.sattar@griffith.edu.au}
\newcommand{\sumaiyaE}{siqbal1@uno.edu}

\newcommand{\usp}{School of Computing, Information and Mathematical Sciences, University of the South Pacific, Laucala Bay, Suva, Fiji}
\newcommand{\umd}{Department of Computer and Information Science, University of Massachusetts Dartmouth, MA, USA}
\newcommand{\uno}{Department of Computer Science, University of New Orleans, LA, USA}
\newcommand{\gu}{Institute for Integrated and Intelligent Systems, Griffith University, Brisbane, QLD, Australia}

\newcommand{\rowSpace}{\renewcommand{\arraystretch}{1.5}}

\newcommand{\inParam}{{\bf INPUT:~}}
\newcommand{\outParam}{{\bf OUTPUT:~}}
\newcommand{\variables}{{\bf VARIABLE:~}}

\newcommand{\myTitle}{
\huge
Guided macro-mutation in a graded energy based genetic algorithm for protein structure prediction\\~
}

\begin{document}
\onecolumn
\begin{frontmatter}
\title{\myTitle\vspace{-1ex}}


\author[SCIMS,GU]{\rashid\corref{mycorrespondingauthor}}
\cortext[mycorrespondingauthor]{Corresponding author}
\ead{\rashidE}
\author[UNO]{\sumaiya}
\ead{\sumaiyaE}
\author[UMD]{\firas}
\ead{\firasE}
\author[UNO]{\tamjid}
\ead{\tamjidE}
\author[GU]{\abdul}
\ead{\abdulE}

\address[SCIMS]{\usp} 
\address[UNO]{\uno} 
\address[UMD]{\umd} 
\address[GU]{\gu} 

\begin{abstract}
\textsf{Protein structure prediction is considered as one of the most challenging and computationally intractable combinatorial problem. Thus, the efficient modeling of convoluted search space, the clever use of energy functions, and more importantly, the use of effective sampling algorithms become crucial to address this problem. For protein structure modeling, an off-lattice model provides limited scopes to exercise and evaluate the algorithmic developments due to its astronomically large set of data-points. In contrast, an on-lattice model widens the scopes and permits studying the relatively larger proteins because of its finite set of data-points. In this work, we took the full advantage of an on-lattice model by using a face-centered-cube lattice that has the highest packing density with the maximum degree of freedom. We proposed a graded energy---strategically mixes the Miyazawa-Jernigan (MJ) energy with the hydrophobic-polar (HP) energy---based genetic algorithm (GA) for conformational search. In our application, we introduced a $2\times2$ HP energy guided macro-mutation operator within the GA to explore the best possible local changes exhaustively. Conversely, the $20 \times 20$ MJ energy model---the ultimate objective function of our GA that needs to be minimized---considers the impacts amongst the 20 different amino acids and allow searching the globally acceptable conformations. On a set of benchmark proteins, our proposed approach outperformed state-of-the-art approaches in terms of the free energy levels and the root-mean-square deviations.}
\end{abstract}
\begin{keyword}
{{\Abi} protein structure prediction; Genetic algorithms; FCC lattice; Miyazawa-Jernigan model; Hydrophobic-Polar model\vspace{-1ex}}
\end{keyword}

\end{frontmatter}


\section{Introduction}
\label{secIntroduction}
Protein folding, by which the primary protein chain with amino acid residue sequence folds into its characteristics and functional three-dimensional (3D) structure in nature, is yet a very complex physical process to simulate~\cite{Morowitz1968Energy_1, Stouthamer1973Weight_2, Alberts2002Cell_3}. Once the folded 3D shape is available, it enables protein to perform specific tasks for living organisms. Conversely, misfolded proteins are responsible for various fatal diseases, such as prion disease, Alzheimer’s disease, Huntington’s disease, Parkinson’s disease, diabetes, and cancer~\cite{Smith2003NatureEd_4, Dobson2003Misfold_5}. Because of these, protein structure prediction (PSP) problem has emerged as a very important research problem.

Homology modeling, threading and \emph{ab initio} are the broad categories of available computational approaches. However, while homologous template is not available, \emph{ab initio} becomes the only computation approach, which aims to find the three dimensional structure of a protein from its primary amino acid sequence alone such that the total interaction energy among the amino acids is minimized.

\emph{Ab initio} computational approach for PSP is a daunting task~\cite{Dodson2007Nature_6} and for modeling the structure on a realistic continuum space such as off-lattice space is even more daunting. However, there are several existing off-lattice models such as Rosetta~\cite{kaufmann2010practically_7}, Quark~\cite{xu2012ab_8}, I-TASSER~\cite{lee2009ab_9}, and so on which map the structures on the realistic continuum spaces rather than using discretized on-lattice spaces and hence, those approaches need to deal with the astronomical data-points incurring heavy computational cost. On-lattice model on the other hand, \emph{i}) due to reduced complexity helps fast algorithms developments and \emph{ii}) widens the scope as well as permits relatively longer protein chains to examine, which is otherwise prohibitive~\cite{Miyazawa1985T20_10, Berrera2003T20_11, Lau1989LatticeHP_12, Cooper2010Foldit_13, das2008macromolecular_14, wroe2005comparing_15}. The computed on-lattice fold can be translated to off-lattice space via hierarchical approaches to provide output in real-space~\cite{hoque2005new_16, hoque2011twin_17, hoque2010dfs_18, Iqbal2015380_19}. The Monte Carlo (MC) or, Conformational Space Annealing (CSA) used in Rosetta can be replaced with better algorithm developed using on-lattice models~\cite{hoque2005new_16, hoque2011twin_17, hoque2010dfs_18}. For instance, we embedded one of our previous on-lattice algorithm~\cite{hoque2011twin_17} within Rosetta and the embedded algorithm improved~\cite{higgs2012applying_21} the average RMSD by 9.5\% and average TM-Score by 17.36\% over the core Rosetta~\cite{kaufmann2010practically_7}. Similarly, the embedded algorithm also outperformed~\cite{higgs2012applying_21} I-TASSER~\cite{lee2009ab_9}. These improvements motivated us further developing superior algorithms using on-lattice models.

The two most important building blocks of an \emph{ab initio} PSP are \textit{i)} an accurate (computable) energy function~\cite{Iqbal2015380_19} and \textit{ii)} an effective search or sampling algorithm. For a simplified model based PSP, it is possible to compute the lower bound~\cite{giaquinta2013effective_20}. It is also possible to know what would be the best score and hence the native score of a sequence by exhaustive enumeration~\cite{unger1993genetic_22, Lesh2003MoveSet_73} (which is feasible to compute for smaller sequences only). Even though, there exists no efficient sampling algorithm yet that can conveniently obtain the known final structure starting from a random structure for all possible available cases~\cite{Hoque2005Guided2D_23, Hoque2007FCC3D_24}. Therefore, a number of efforts are being made, such as, different types of meta-heuristics have been used in solving the on lattice PSP problems. These include Monte Carlo Simulation~\cite{Thachuk2007ReplicaMC_25}, Simulated Annealing~\cite{Tantar2008SA_26}, Genetic Algorithms (GA)~\cite{Hoque2005Guided2D_23, Hoque2007FCC3D_24, Unger1993Genetic3D_27, Hoque2007PhDThesis_28}, Tabu Search with GA~\cite{Bockenhauer2008Tabu_29}, GA with twin-removal operator~\cite{Hoque2011Twin_30}, Tabu Search with Hill Climbing~\cite{Klau2002HumanGuided_31}, Ant Colony Optimization~\cite{Blum2005Ant_32}, Particle Swarm Optimization~\cite{Kondov2011PSO_33, Mansour2012PSO_34}, Immune Algorithms~\cite{Cutello2007AIS_35}, Tabu-based Stochastic Local Search~\cite{Cebrian2008TabuFcc_36, Shatabda2012Mem_37}, Firefly Algorithm~\cite{maher2014firefly_38}, and Constraint Programming~\cite{Mann2008CPSP_39, Dotu2011OnLattice_40}.

Krasnogor et al.~\cite{krasnogor2002multimeme_41} applied HP model for PSP problem using the square, triangular, and diamond lattices and further extended their work applying fuzzy-logic~\cite{pelta2005multimeme_42}. Islam et al. further improved the performance of memetic algorithms in a series of work~\cite{kamrul2011Phd_43, IslamC09_44, IslamCUS11_45, IslamCM11_46} for the simplified PSP models. They also proposed a clustered architecture for the memetic algorithm with a scalable niching technique~\cite{IslamC10_47, IslamCM11_48, islam2013clustered_49} for PSP. However, using 3D FCC lattice points, the recent state-of-the-art results for the HP energy model have been achieved by genetic algorithms~\cite{Rashid2012GAPlus_50, Swakkhar2013Encode_51}, local search approaches~\cite{Shatabda2012Mem_37, Rashid2013SS_52}, a local search embedded GA~\cite{Rashid2013LSEGA_53}, and a multi-point parallel local search approach~\cite{Rashid2013PSS_54}. Kern and Lio~\cite{kern2013lattice_75} applied hydrophobic-core guided genetic operator for efficient searching on HP, HPNX and hHPNX lattice models. Several approaches towards the $20 \times 20$ energy model include a constraint programming technique used in~\cite{Palu2004Constraint_55, Palu2005Heuristics_56} by to predict tertiary structures of real proteins using secondary structure information, a fragment assembly method~\cite{Palu2011CLP_57} to optimize protein structures. Among other successful approaches, a population based local search~\cite{Kapsokalivas2009Population_58} and a population based genetic algorithm~\cite{Torres2007GAT20_59} are found in the literature that applied empirical energy functions.

In a hybrid approach, Ullah et al.~\cite{Ullah2010Hybrid_60} applied a constraint programming based large neighborhood search technique on top of the output of COLA~\cite{Palu2007COLA_61} solver. The hybrid approach produced the state-of-the-art results for several small sized (less than 75 amino acids) benchmark proteins. In another work, Ullah et al.~\cite{Ullah2009TwoStage_62} proposed a two stage optimization approach combining constraint programming and local search using Berrera {\etal}~\cite{Berrera2003T20_11} deduced {\ttx} energy matrix (we denote this model as BM). In a recent work~\cite{Shatabda2013Heuristic_63}, Shatabda et al. presented a mixed heuristic local search algorithm for PSP and produced the state-of-the-art results using the BM model on 3D FCC lattice. Although the heuristics themselves are weaker than the BM energy model, their collective use in the random mixing fashion produce results better than the BM energy itself. In a previous work \cite{Rashid2013Mixed}, we applied BM and HP energy models in a mixed manner within a GA framework and showed that hybridizing energies performs better than their individual performances.

In this work, we propose a graded as well as hybrid energy function with a genetic algorithm (GA) based sampling to develop an effective \emph{ab initio} PSP tool. The graded energy-model strategically mixes $20 \times 20$ Miyazawa-Jernigan (MJ) contact-energy~\cite{Miyazawa1985T20_10, Berrera2003T20_11} with the simple $2 \times 2$ Hydrophobic-Polar (HP) contact-energy model~\cite{Lau1989LatticeHP_12}, denoted as MH (MJ+HP $\rightarrow$ MH) in this paper. Specifically, we propose a hydrophobic-polar categorization of the HP model within a hydrophobic-core directed macro-mutation operator to explore the local benefits exhaustively while the GA sampling is guided by the MJ energy Matrix globally. While the fine grained details of the high resolution interaction energy matrix can become computationally prohibit, a low resolution energy model may effectively sample the search-space towards certain promising directions particularly emphasizing on the pair-wise contributions with large magnitudes – which we have implementation strategically via a macro mutation. Further, we use an enhanced genetic algorithm (GA) framework~\cite{Rashid2012GAPlus_50} for protein structure optimization on 3D face-centered-cube (FCC) lattice model. Prediction in the FCC lattice model can yield the densest protein core~\cite{Hoque2005Guided2D_23} and the FCC lattice model can provide the maximum degree of freedom as well as the closest resemblance to the real or, high resolution folding within the lattice constraint. FCC orientation can therefore align a real protein into the closest conformation amongst the available lattice configurations~\cite{Hoque2007FCC3D_24}.

On a set of standard benchmark proteins, our MH model guided GA, named as \emph{MH\_GeneticAlgorithm} (MH\_GA), shows significant improvements in terms of interaction energies and root-mean-square deviations in comparison to the state-of-the-art search approaches~\cite{Ullah2010Hybrid_60, Shatabda2013Heuristic_63, Torres2007GAT20_59} for the lattice based PSP models. For a fair comparison, we run~\cite{Ullah2010Hybrid_60} and~\cite{Shatabda2013Heuristic_63} using MJ energy model and in the result section, we compare our experimental results with the results produced by~\cite{Ullah2010Hybrid_60} and~\cite{Shatabda2013Heuristic_63}. Further, we present an experimental analysis showing the effectiveness of using the hydrophobic polar categorization of the HP model to direct macro-mutation operation.

\section{Background}
\label{secBackground}
Anfinsen's hypothesis~\cite{Anfinsen1973GovernFolding_69} and Levinthal's paradox~\cite{Levinthal1968Pathway_68} form the basis and the confidence of the {\abi} approach, which inform that the protein structure prediction can be relied only on the amino acid sequence of the target protein as well as there should be a non-exhaustive pathway to obtain the native fold. Thus, we set our goal to model the folding process using on-lattice model. Further, it has been argued in~\cite{alm1999matching_64, baker2000surprising_65}, ``... protein folding mechanisms and landscapes are largely determined by the topology of the native state and are relatively insensitive to details of the interatomic interactions. This dependence on low resolution structural features, rather than on high resolution detail, suggests that it should be possible to describe the fundamental physics of the folding process using relatively low resolution models ... The observation that protein folding mechanisms are determined primarily by low resolution topological features and not by high resolution details suggests that a simple theory incorporating features of the native state topology should be successful in predicting the broad outlines of folding reactions''. A rigorous discussion on on-lattice models can be found in~\cite{istrail2009combinatorial_66, bahi2013computational_67}. Further, exploration of an astronomically large search space and the evaluation of the conformations using a real energy models are the big challenges often being computationally prohibitive, especially for sequences $>$ 200 residues long, whereas simplified models can aid in modeling and understanding the protein folding process feasibly. Next, we describe the model that we use.

\subsection{Simplified model}
In our simplified model, we use 3D FCC lattice points to map the amino acids of a protein sequence. In the mapping, each amino acid of the sequence, occupies a point on the lattice to form a continuous chain of a self-avoiding-walk. We apply the MJ energy matrix in conjunction with the HP energy model in a genetic algorithm framework for PSP. The FCC lattice, the HP and MJ energy models, and the GA are briefly described below.

\begin{figure}[!h]
\centering
	\includegraphics[height=10cm]{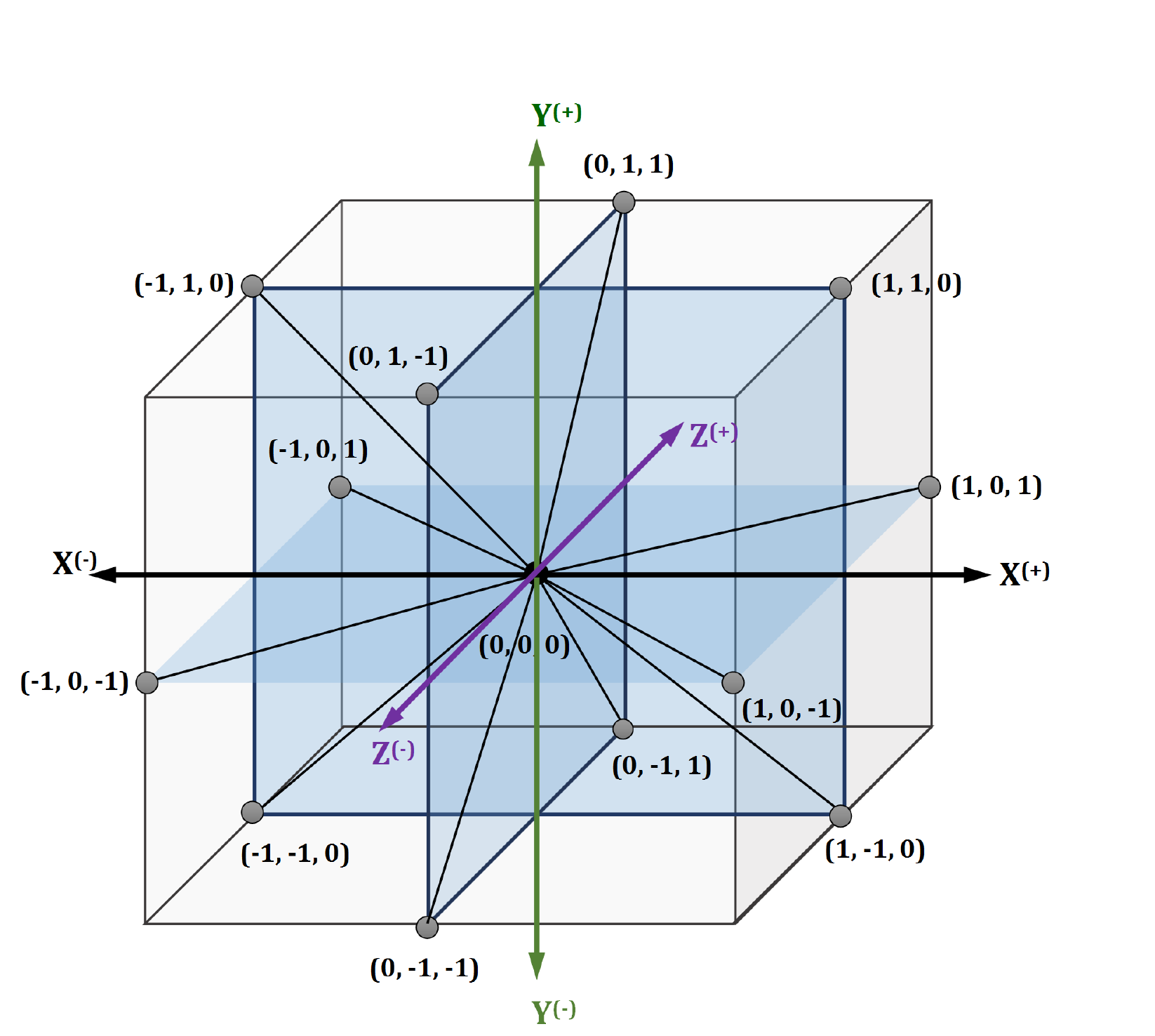}
	\caption{\small {A 3-dimensional {\fcc} lattice space.} The $12$ basis vectors of the neighbors of the origin $(0,0,0)$ in a Cartesian coordinate system.}
	\label{simpleFCC}
\end{figure}

\subsubsection*{FCC lattice}
The FCC lattice has the highest packing density compared to the other existing lattices~\cite{Hales2005Proof_70}. Thus, FCC model can provide maximum degree of freedom within a constrained space. In FCC, each lattice point (the origin in Figure 1) has 12 neighbors with closest possible distance having 12 \textit{basis vectors} as follows:

\begin{center}
	\setlength{\tabcolsep}{5pt}
	\rowSpace
	\begin{footnotesize}
	\begin{tabular}{llll}
		{$v_1=(1,1,0)$}&{$v_4=(-1,-1,0)$}&{$v_7=(-1,1,0)$}&{$v_{10}=(0,1,-1)$}\\
		{$v_2=(1,0,1)$}&{$v_5=(-1,0,-1)$}&{$v_8=(1,-1,0)$}&{$v_{11}=(1,0,-1)$}\\
		{$v_3=(0,1,1)$}&{$v_6=(0,-1,-1)$}&{$v_9=(-1,0,1)$}&{$v_{12}=(0,-1,1)$}
	\end{tabular}
	\end{footnotesize}
\end{center}

In simplified PSP, conformations are mapped on the lattice by a sequence of basis vectors, or by the {\em relative vectors} that are relative to the previous basis vectors in the sequence.

\subsubsection*{HP energy model}
\begin{figure}[!h]
\centering
\begin{tabular}{ccc}
\includegraphics[width=4cm]{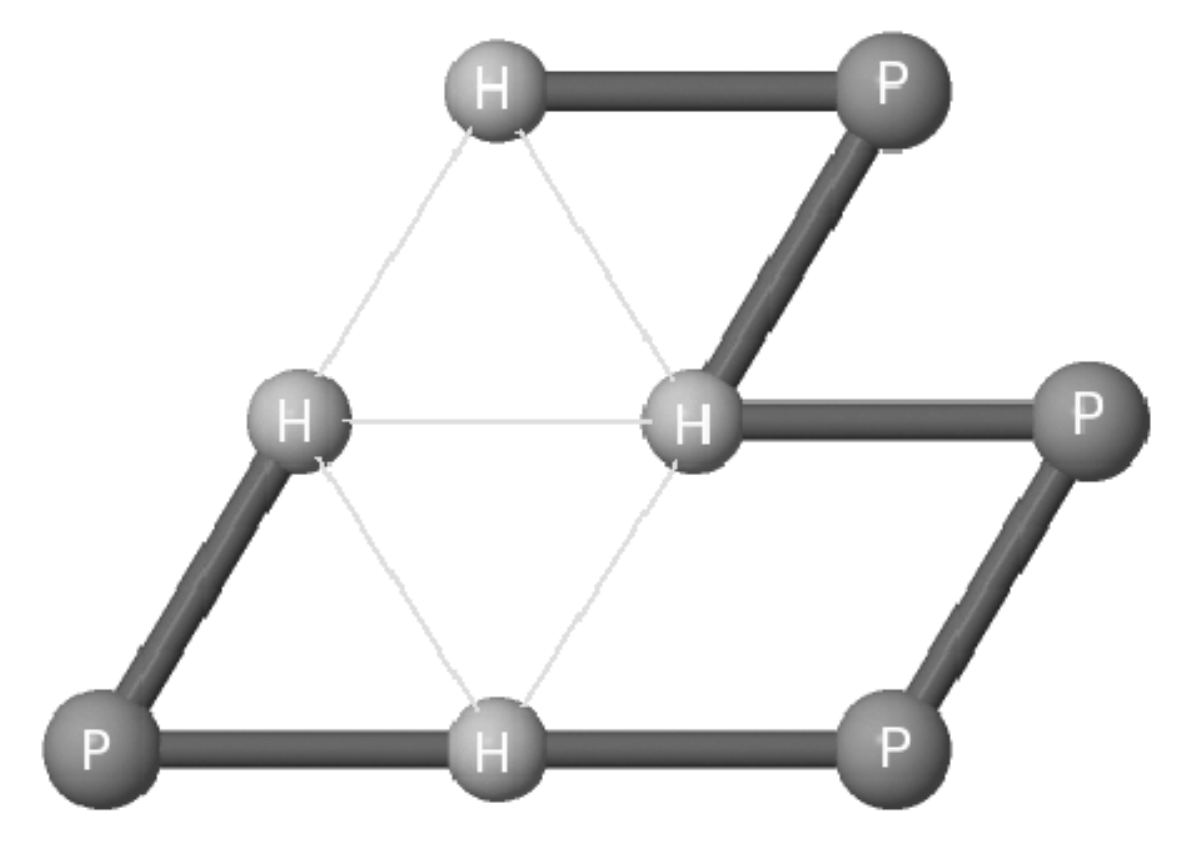}&&
\includegraphics[width=4cm]{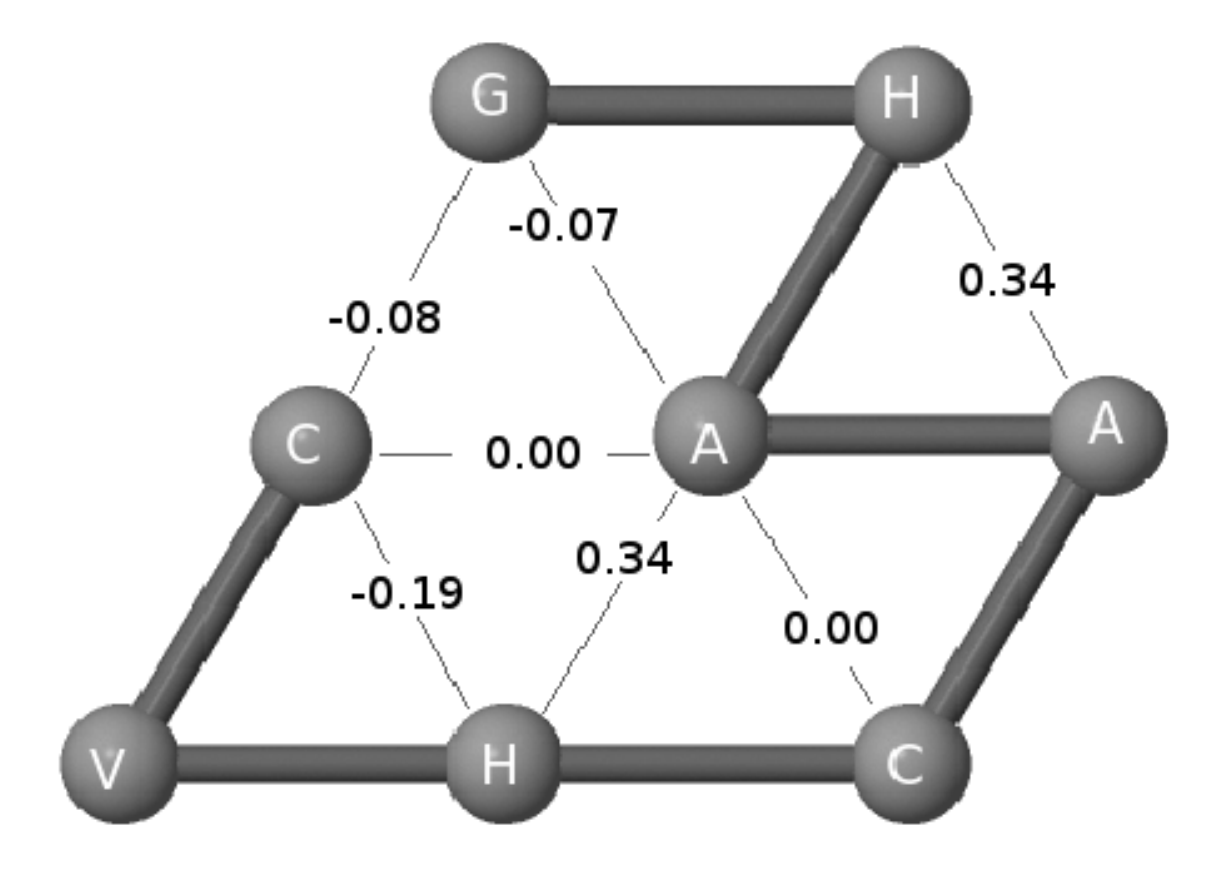}\\
{a) HP energy calculation}&~~~~~~&{b) MJ energy calculation}
\end{tabular}
\caption{\small On a lattice based model, (a) is showing H-H contact energy $-4$ ($-1\times4$) using the HP model for a random sequence: \emph{HPHPPHPH} (b) is showing the sum of the pair-wise contact potentials $0.34$ ($(-0.08)+(-0.19)+(0.00)\times2+(-0.07)+(0.34)\times2$) using the MJ model for a random sequence: \emph{GHAACHVC}.}
\label{energy_calculate}
\end{figure}

Based on the hydrophobic property, the $20$ {\aas} which are the constituents of all proteins, are broadly divided into two categories: (a) hydrophobic {\aas} (\emph{Gly}, \emph{Ala}, \emph{Pro}, \emph{Val}, \emph{Leu}, \emph{Ile}, \emph{Met}, \emph{Phe}, \emph{Tyr}, \emph{Trp}) are denoted as H; and (b) hydrophilic or polar {\aas}  (\emph{Ser}, \emph{Thr}, \emph{Cys}, \emph{Asn}, \emph{Gln}, \emph{Lys}, \emph{His}, \emph{Arg}, \emph{Asp}, \emph{Glu}) are denoted as P. In the $2 \times 2$ HP model~\cite{Lau1989LatticeHP_12}, when two non-consecutive hydrophobic amino acids become topologically neighbors, they contribute a certain amount of negative energy, which for simplicity is considered as $-1$ (Table~\ref{hp_matrix}). The total energy $E_{HP}$ (Equation \ref{eqHP}) of a conformation based on the HP model becomes the sum of the contributions over all pairs of the non-consecutive hydrophobic amino acids ({\fig}\ref{energy_calculate}a).

\begin{table}[h]
\caption{\small The $2 \times 2$ HP energy model.}
\label{hp_matrix}
\centering
\vspace{1ex}
\begingroup
\fontsize{10pt}{20pt}\selectfont
\fontsize{10pt}{\baselineskip}
	\setlength{\tabcolsep}{10pt}
	\renewcommand{\arraystretch}{1}
	\begin{tabular}{|l|rr|}
	\hline
	{}&{H  }&{ P  }\\
	\hline
	{H}&{-1}&{0}\\
	{P}&{0}&{0}\\
	\hline
\end{tabular}
\endgroup
\end{table}

\begin{equation}
	 E_\mathrm{HP}=\sum_{i<j-1} \mathsf{c}_{ij} \times \mathsf{e}_{ij}
	\label{eqHP}
\end{equation}

where, $\mathsf{c}_{ij}=1$ if amino acids at positions $i$ and $j$ in the sequence are non-consecutive but topological neighbors on the lattice, otherwise $c_{ij}$ = $0$. The $\mathsf{e}_{ij}=-1$ if the $i\mathsf{th}$ and $j\mathsf{th}$ amino acids are both hydrophobic, otherwise $e_{ij}$ = $0$.

\begin{table}[h]
\caption{\small The {\ttx} MJ energy model by Miyazawa and Jernigan~\cite{Miyazawa1985T20_10}. The bold-faced amino acids are Hydrophobic (H) and others are Polar (P).}
\label{mj_matrix}
\centering
\vspace{1ex}
	\setlength{\tabcolsep}{1.5pt}
	\renewcommand{\arraystretch}{1.2}
	\begin{scriptsize}

	\begin{tabular}{|l|rrrrrrrrrrrrrrrrrrrr|}
	\hline
	{CYS}&{-1.06}&{}&{}&{}&{}&{}&{}&{}&{}&{}&{}&{}&{}&{}&{}&{}&{}&{}&{}&{}\\
	{\bf MET}&{0.19}&{0.04}&{}&{}&{}&{}&{}&{}&{}&{}&{}&{}&{}&{}&{}&{}&{}&{}&{}&{}\\
	{\bf PHE}&{-0.23}&{-0.42}&{-0.44}&{}&{}&{}&{}&{}&{}&{}&{}&{}&{}&{}&{}&{}&{}&{}&{}&{}\\
	{\bf ILE}&{0.16}&{-0.28}&{-0.19}&{-0.22}&{}&{}&{}&{}&{}&{}&{}&{}&{}&{}&{}&{}&{}&{}&{}&{}\\
	{\bf LEU}&{-0.08}&{-0.20}&{-0.30}&{-0.41}&{-0.27}&{}&{}&{}&{}&{}&{}&{}&{}&{}&{}&{}&{}&{}&{}&{}\\
	{\bf VAL}&{0.06}&{-0.14}&{-0.22}&{-0.25}&{-0.29}&{-0.29}&{}&{}&{}&{}&{}&{}&{}&{}&{}&{}&{}&{}&{}&{}\\
	{\bf TRP}&{0.08}&{-0.67}&{-0.16}&{0.02}&{-0.09}&{-0.17}&{-0.12}&{}&{}&{}&{}&{}&{}&{}&{}&{}&{}&{}&{}&{}\\
	{\bf TYR}&{0.04}&{-0.13}&{0.00}&{0.11}&{0.24}&{0.02}&{-0.04}&{-0.06}&{}&{}&{}&{}&{}&{}&{}&{}&{}&{}&{}&{}\\
	{\bf ALA}&{0.00}&{0.25}&{0.03}&{-0.22}&{-0.01}&{-0.10}&{-0.09}&{0.09}&{-0.13}&{}&{}&{}&{}&{}&{}&{}&{}&{}&{}&{}\\
	{\bf GLY}&{-0.08}&{0.19}&{0.38}&{0.25}&{0.23}&{0.16}&{0.18}&{0.14}&{-0.07}&{-0.38}&{}&{}&{}&{}&{}&{}&{}&{}&{}&{}\\
	{THR}&{0.19}&{0.19}&{0.31}&{0.14}&{0.20}&{0.25}&{0.22}&{0.13}&{-0.09}&{-0.26}&{0.03}&{}&{}&{}&{}&{}&{}&{}&{}&{}\\
	{SER}&{-0.02}&{0.14}&{0.29}&{0.21}&{0.25}&{0.18}&{0.34}&{0.09}&{-0.06}&{-0.16}&{-0.08}&{0.20}&{}&{}&{}&{}&{}&{}&{}&{}\\
	{GLN}&{0.05}&{0.46}&{0.49}&{0.36}&{0.26}&{0.24}&{0.08}&{-0.20}&{0.08}&{-0.06}&{-0.14}&{-0.14}&{0.29}&{}&{}&{}&{}&{}&{}&{}\\
	{ASN}&{0.13}&{0.08}&{0.18}&{0.53}&{0.30}&{0.50}&{0.06}&{-0.20}&{0.28}&{-0.14}&{-0.11}&{-0.14}&{-0.25}&{-0.53}&{}&{}&{}&{}&{}&{}\\
	{GLU}&{0.69}&{0.44}&{0.27}&{0.35}&{0.43}&{0.34}&{0.29}&{-0.10}&{0.26}&{0.25}&{0.00}&{-0.26}&{-0.17}&{-0.32}&{-0.03}&{}&{}&{}&{}&{}\\
	{ASP}&{0.03}&{0.65}&{0.39}&{0.59}&{0.67}&{0.58}&{0.24}&{0.00}&{0.12}&{-0.22}&{-0.29}&{-0.31}&{-0.17}&{-0.30}&{-0.15}&{0.04}&{}&{}&{}&{}\\
	{HIS}&{-0.19}&{0.99}&{-0.16}&{0.49}&{0.16}&{0.19}&{-0.12}&{-0.34}&{0.34}&{0.20}&{-0.19}&{-0.05}&{-0.02}&{-0.24}&{-0.45}&{-0.39}&{-0.29}&{}&{}&{}\\
	{ARG}&{0.24}&{0.31}&{0.41}&{0.42}&{0.35}&{0.30}&{-0.16}&{-0.25}&{0.43}&{-0.04}&{-0.35}&{0.17}&{-0.52}&{-0.14}&{-0.74}&{-0.72}&{-0.12}&{0.11}&{}&{}\\		{LYS}&{0.71}&{0.00}&{0.44}&{0.36}&{0.19}&{0.44}&{0.22}&{-0.21}&{0.14}&{0.11}&{-0.09}&{-0.13}&{-0.38}&{-0.33}&{-0.97}&{-0.76}&{0.22}&{0.75}
			&{0.25}&{}\\
	{\bf PRO}&{0.00}&{-0.34}&{0.20}&{0.25}&{0.42}&{0.09}&{-0.28}&{-0.33}&{0.10}&{-0.11}&{-0.07}&{0.01}&{-0.42}&{-0.18}&{-0.10}&{0.04}&{-0.21}&{-0.38}
			&{0.11}&{0.26}\\
	\hline
	{}&{CYS  }&{\bf MET  }&{\bf PHE}&{\bf ILE}&{\bf LEU}&{\bf VAL}&{\bf TRP}&{\bf TYR}&{\bf ALA}&{\bf GLY}
		&{THR}&{SER}&{GLN}&{ASN}&{GLU}&{ASP}&{HIS}&{ARG}&{LYS}&{\bf PRO}\\
	\hline
\end{tabular} 
\end{scriptsize}
\end{table}

\subsubsection*{MJ energy model}
By analyzing crystallized protein structures, Miyazawa and Jernigan~\cite{Miyazawa1985T20_10} statistically deduced a {\ttx} energy matrix (better known as MJ energy model) that considers residue contact propensities between the amino acids. BM is a similar energy matrix as MJ deduced by Berrera {\etal}~\cite{Berrera2003T20_11} by calculating empirical contact energies on the basis of information available from a set of selected protein structures and following the quasi-chemical approximation. In this work, we use MJ energy model. The total energy $E_{MJ}$ (Equation \ref{eqT20}) of a conformation based on the MJ energy model is the sum of the contributions of all of the non-consecutive amino acid pairs that are topological neighbors ({\fig}\ref{energy_calculate}b).

\begin{equation}
	 E_{\mathrm{MJ}}=\sum_{i<j-1} \mathsf{c}_{ij} \times \mathsf{e}_{ij}
	\label{eqT20}
\end{equation}

where, $\mathsf{c}_{ij}=1$ if amino acids at positions $i$ and $j$ in the sequence are non-consecutive neighbors on the lattice, otherwise $c_{ij}$ = $0$; and $\mathsf{e}_{ij}$ is the empirical energy value between the $i\mathsf{th}$ and $j\mathsf{th}$ amino acid pair specified in the MJ energy matrix as shown in {\tab}\ref{mj_matrix}.

\begin{algorithm}[!h]
\footnotesize\sf
	\tcc{{\inParam}Protein sequence, Crossover rate, Mutation rate, Population size}
	\tcc{{\outParam}Global best solution}
	\vspace{1ex}
    initialize population with encoded protein sequences as chromosomes (individuals)\;
    evaluate population\;
  \Repeat{(termination criteria)}{
    Select the best-fit individuals for reproduction\;
    Breed new individuals using crossover and mutation operations according to the rates\;
    Evaluate the new individuals\;
    Replace least-fit individuals with new best-fit individuals\;
   }
  \caption{The standard genetic algorithm}
  \label{algGenericGA}
\end{algorithm}

\subsection{Genetic algorithms}
GAs~\cite{holland1975adaptation_71} are a family of population-based search algorithms which can be applied for PSP as an optimization problem. The outline of GA as given in Algorithm~\ref{algGenericGA}, follows simple steps: Line 1 initializes the population; the Line 2 evaluates the solutions to rank them by relative quality; and the Lines 4-7 are repeating on generating, evaluating and replacing the least-fitted off-springs within the population until the termination criteria arises. For the coding scheme, non-isomorphic encoding~\cite{hoque2006non_72} has been applied and the $v_1, \ldots, v_{12}$ (in Figure~\ref{simpleFCC}) can be thought of be renamed as \emph{a}, $\ldots$, \emph{l} respectively.

\begin{figure}[!h]
\centering
	\includegraphics[width=.6\textwidth]{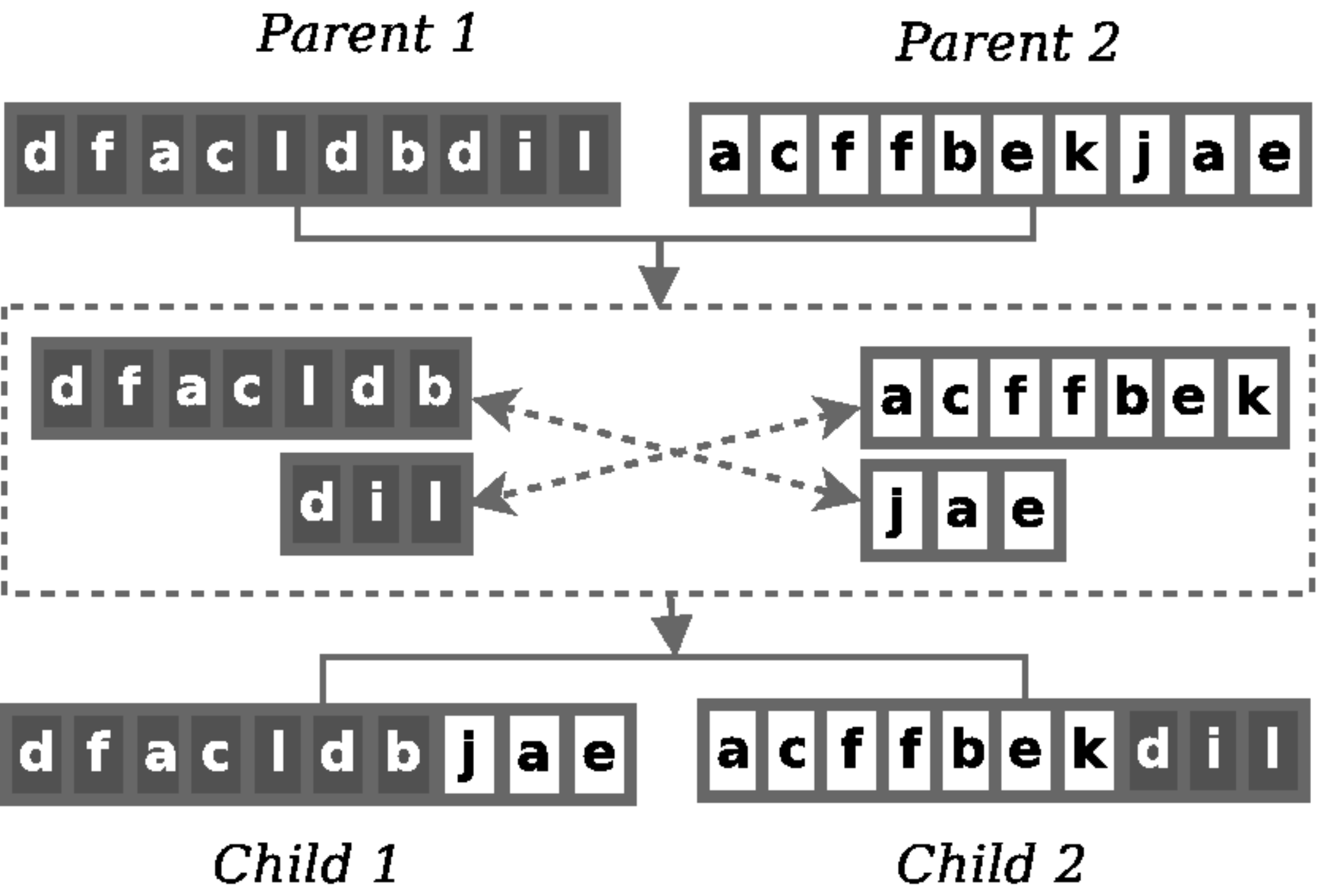}\\
	\caption{\footnotesize Typical crossover operator: exchanging parts and forming new chromosomes.}
	\label{ga-crossover}
\end{figure}

\begin{figure}[h]
\centering
	\includegraphics[width=.6\textwidth]{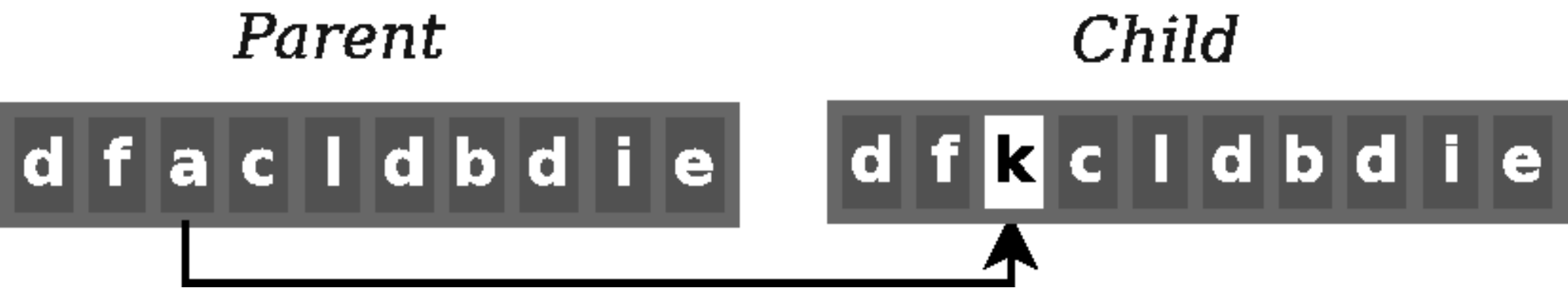}\\
	\caption{\footnotesize Typical mutation operator: mutating one point into some other point.}
	\label{ga-mutation}
\end{figure}

A typical crossover operator randomly splits two solutions at a randomly selected crossover point and exchanges the parts between them ({\fig}\ref{ga-crossover}) and a typical mutation operator alters a solution at a random point ({\fig}\ref{ga-mutation}). In the case of PSP, conformations are regarded as solutions of a GA population.

\section{Methods}
\label{secMethods}
This section describes the proposed MH\_GA framework along with the implementation level detail.  We implemented the framework in Java (J2EE), using Rocks clusters. The code for MH based GA is freely available online\footnote{\scriptsize{Download JAR files from: \url{http://cs.uno.edu/~tamjid/Software/MH_GA/JarFiles.zip}}}. 

\begin{figure*}[!h]
\setlength{\tabcolsep}{0pt}
\renewcommand{\arraystretch}{2}
\centering
\begin{tabular}{ccc|cccc}
	\multicolumn{3}{c|}{\bf Crossover operator}&\multicolumn{4}{c}{\bf Mutation operators}\\
	\hline
\includegraphics[scale=0.12]{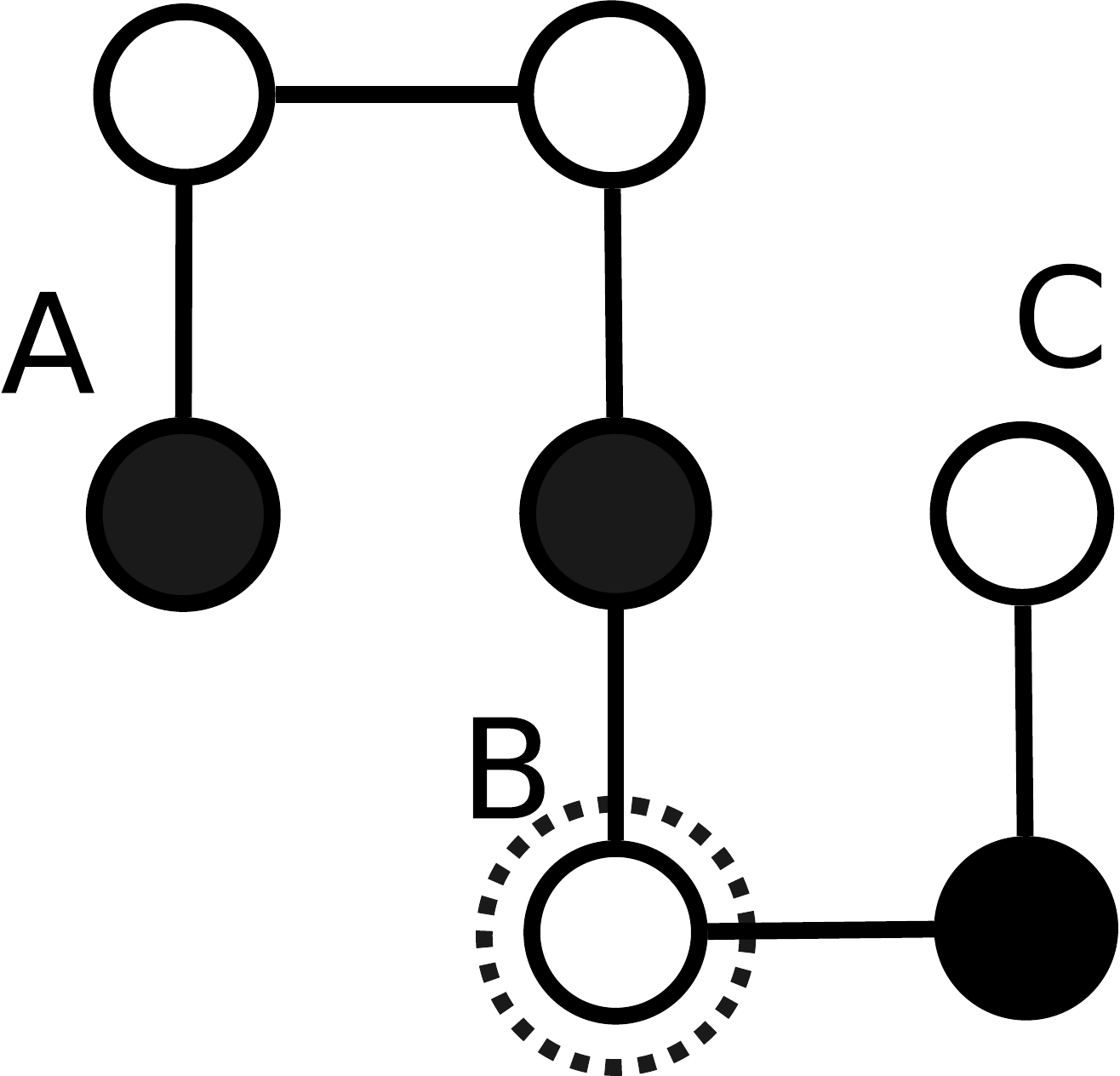}&
&\includegraphics[scale=0.12]{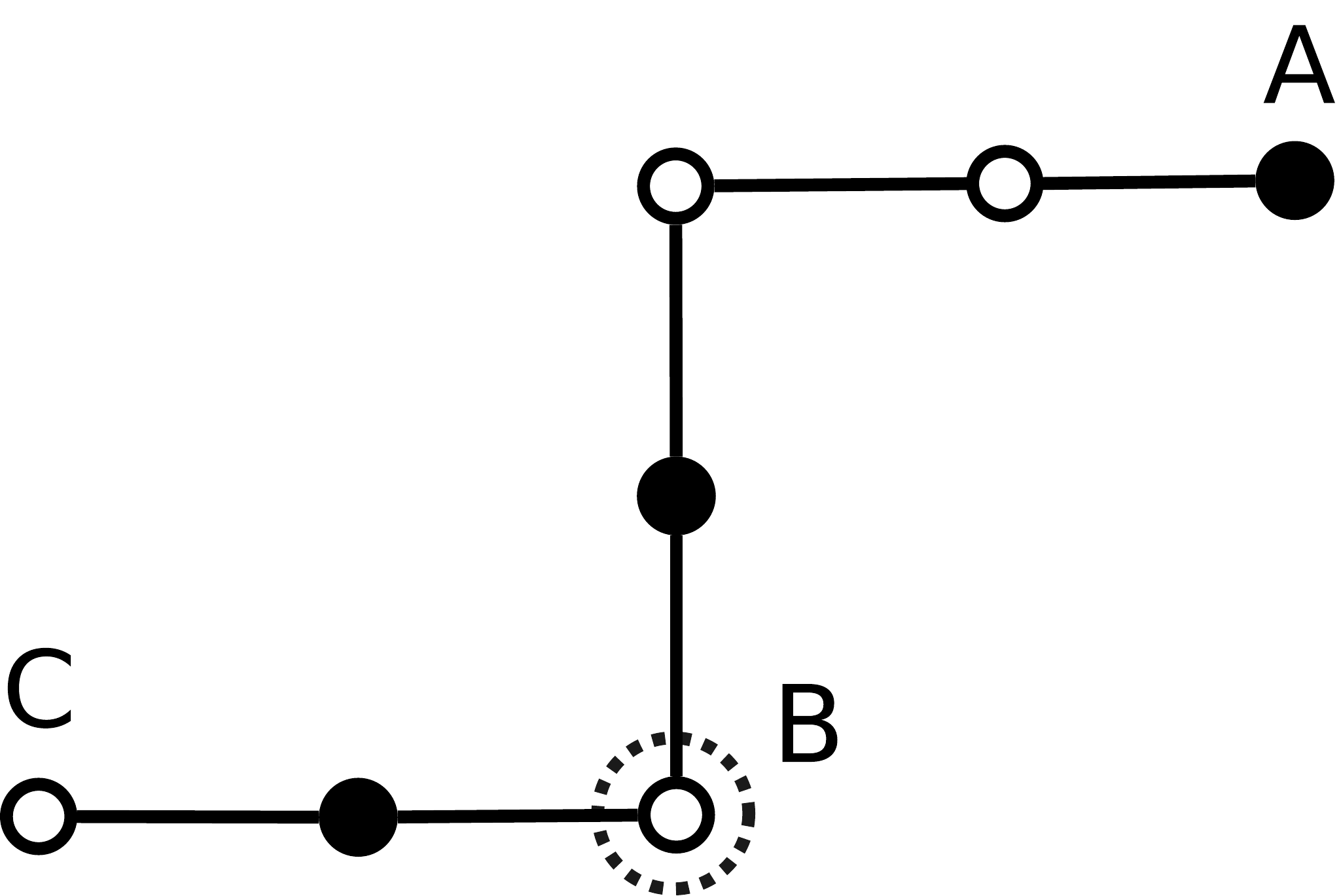}
&\includegraphics[scale=0.12]{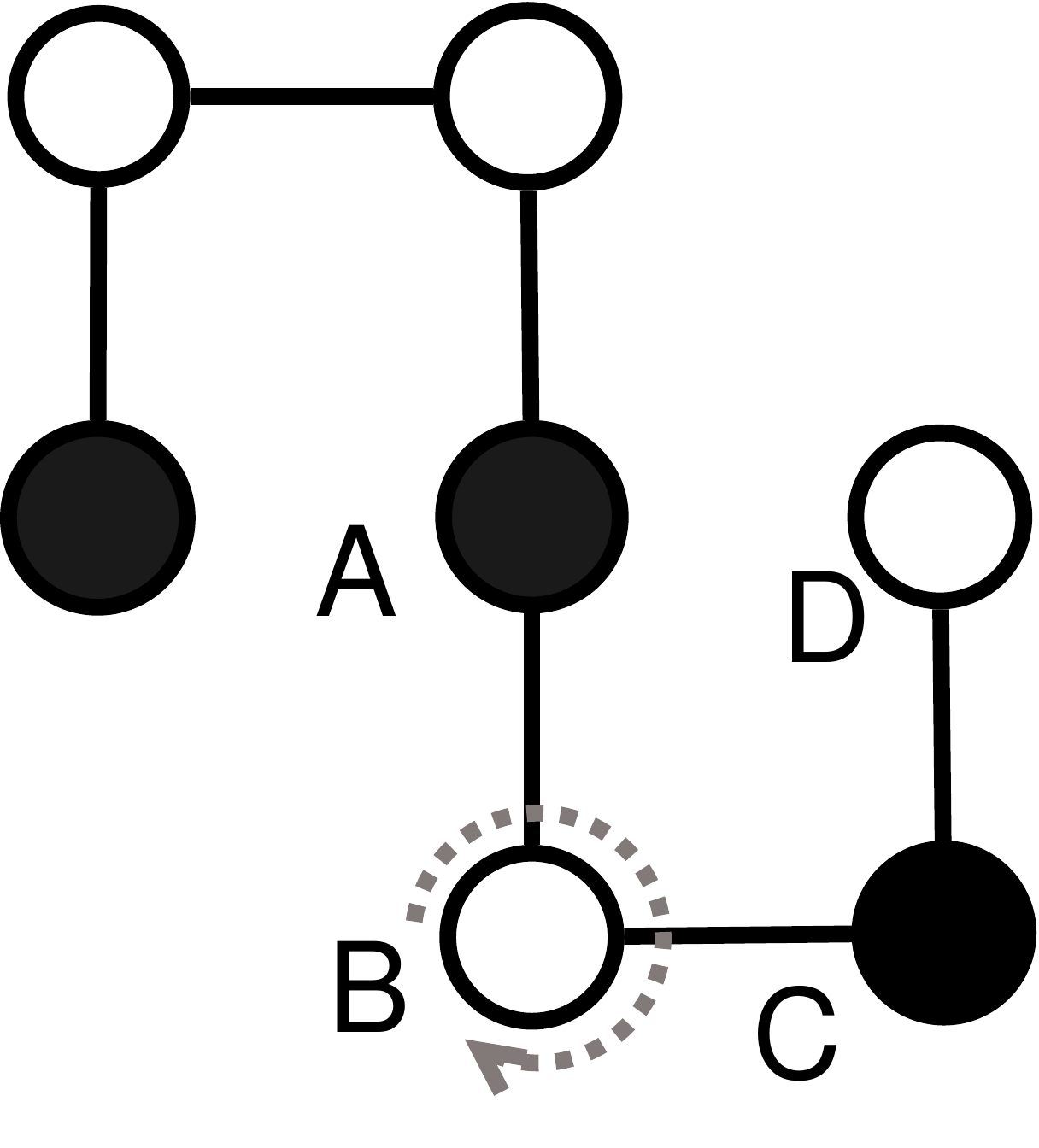}
&\includegraphics[scale=0.12]{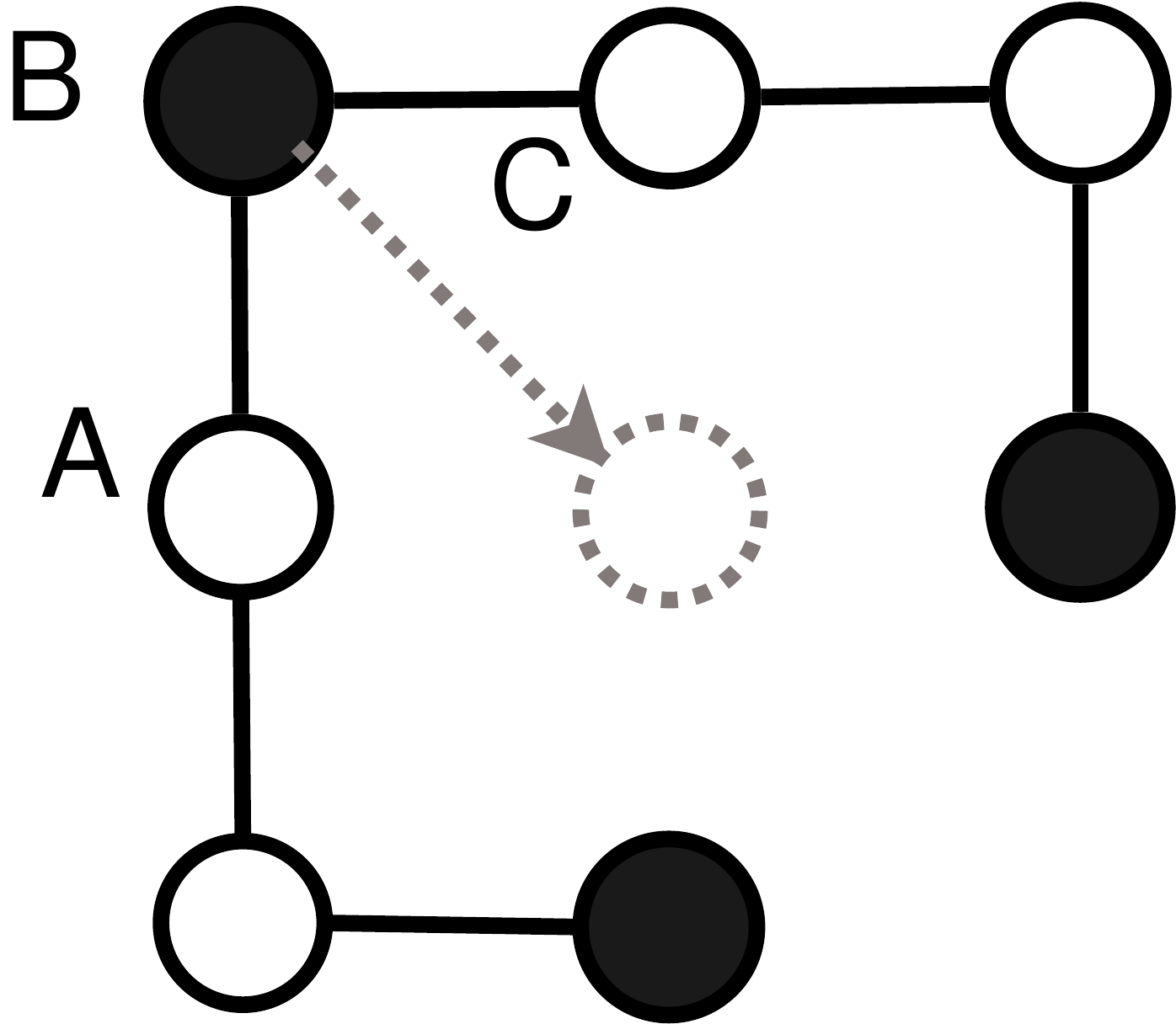}
&\includegraphics[scale=0.12]{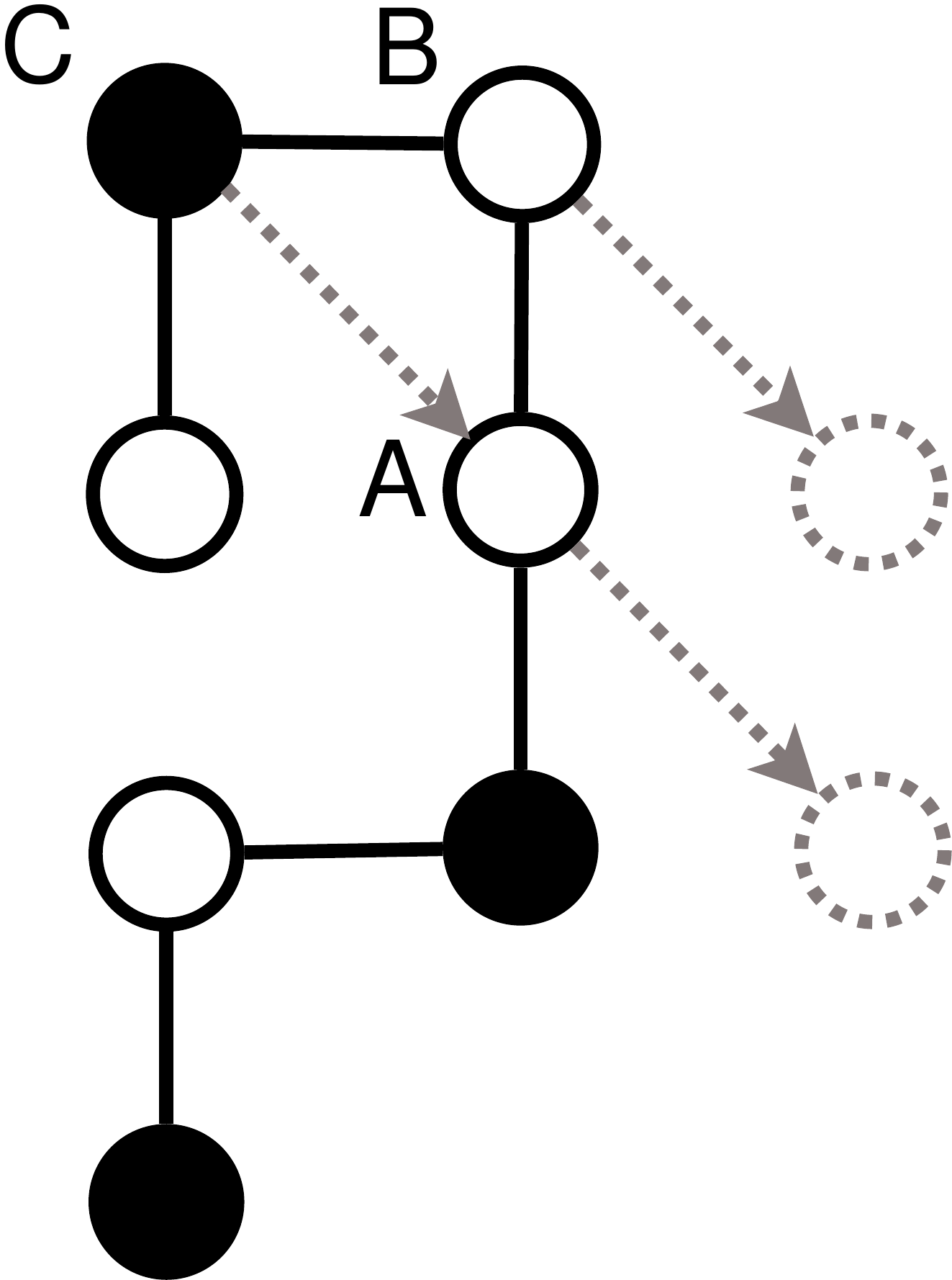}
&\includegraphics[scale=0.12]{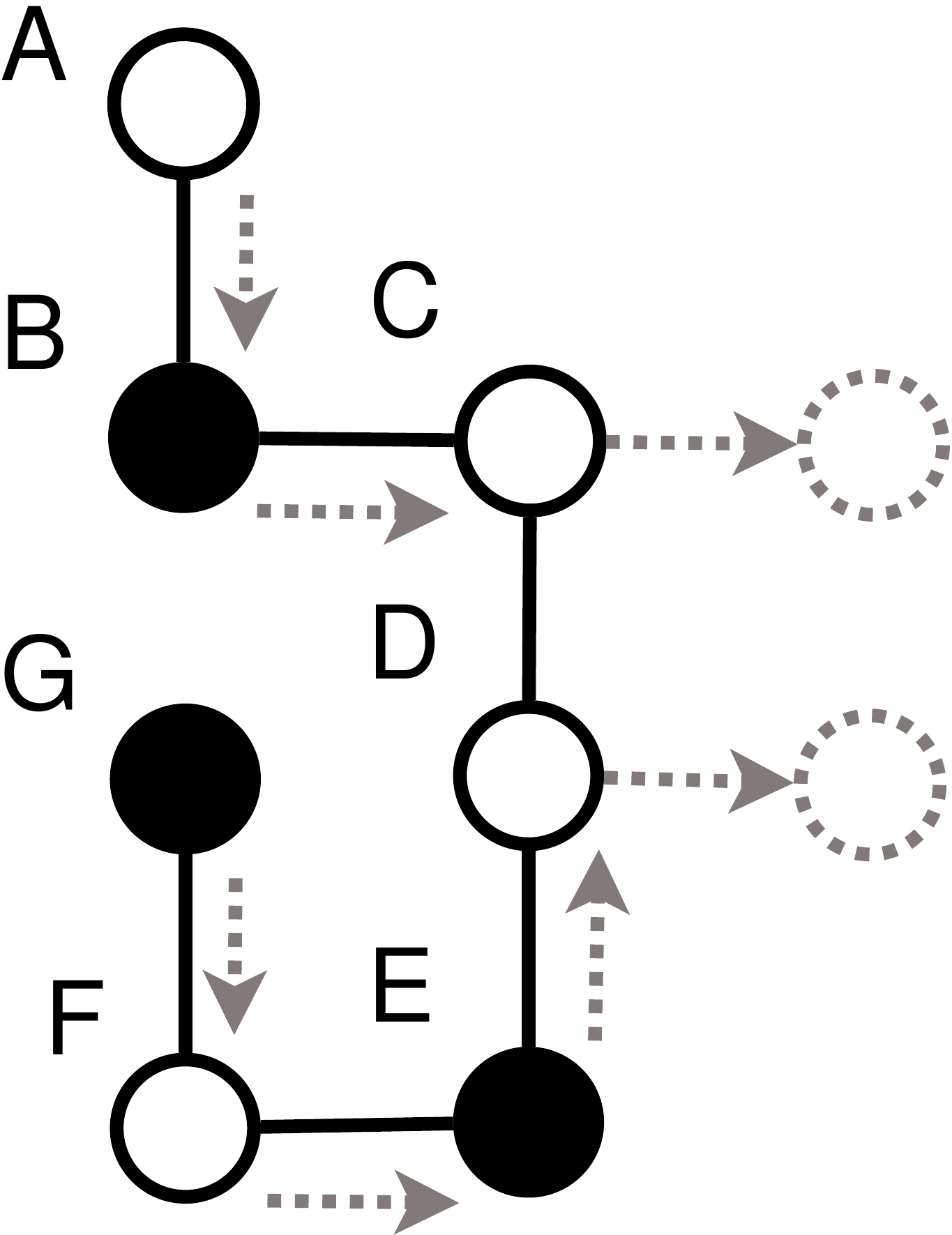}\\
	&\includegraphics[scale=0.10]{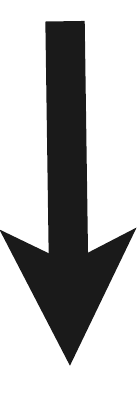}&
	&\includegraphics[scale=0.10]{figure05_arrowd-eps-converted-to.pdf}
		&\includegraphics[scale=0.10]{figure05_arrowd-eps-converted-to.pdf}
		&\includegraphics[scale=0.10]{figure05_arrowd-eps-converted-to.pdf}
		&\includegraphics[scale=0.10]{figure05_arrowd-eps-converted-to.pdf}\\
\includegraphics[scale=0.12]{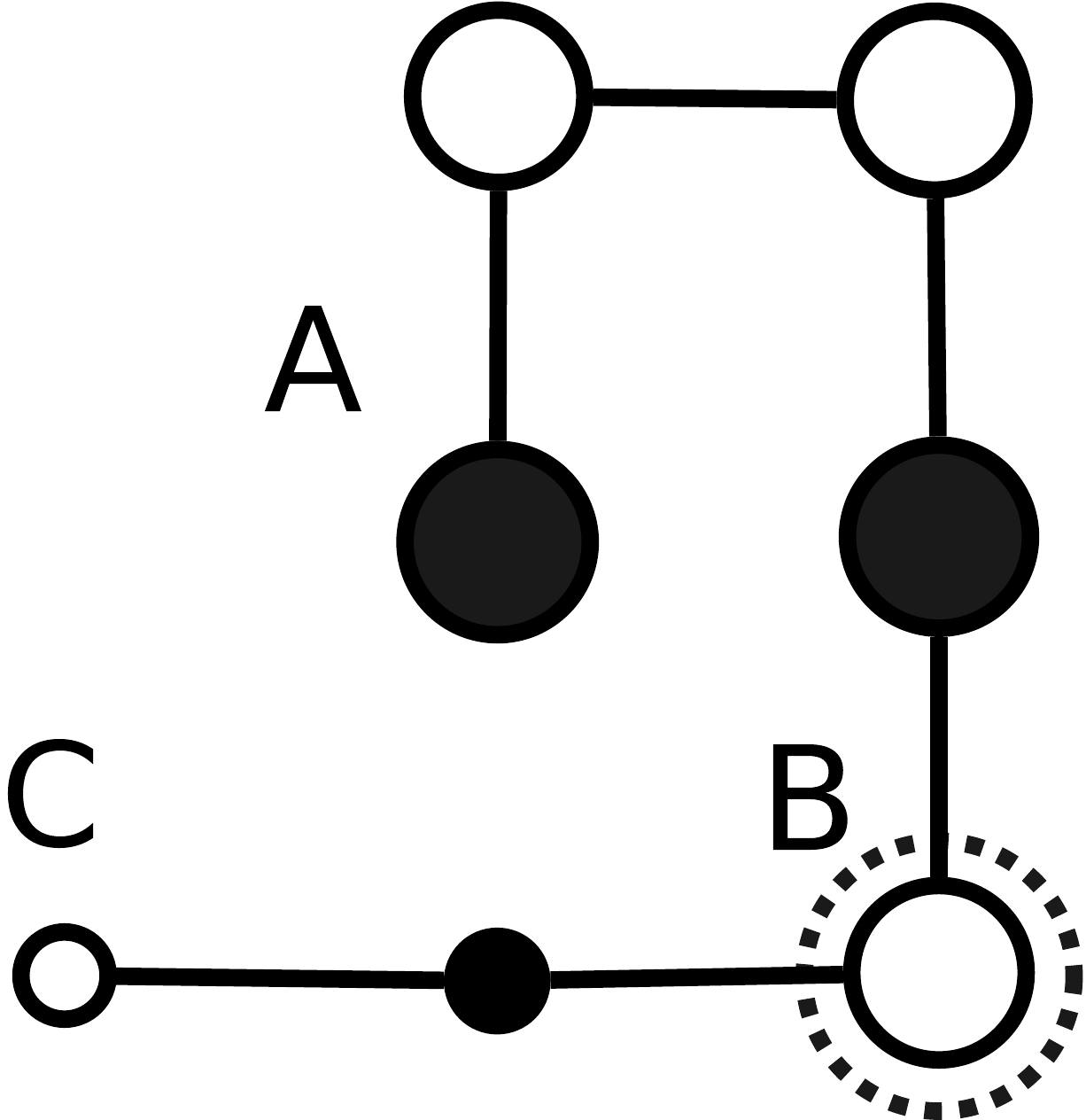}&&\includegraphics[scale=0.15]{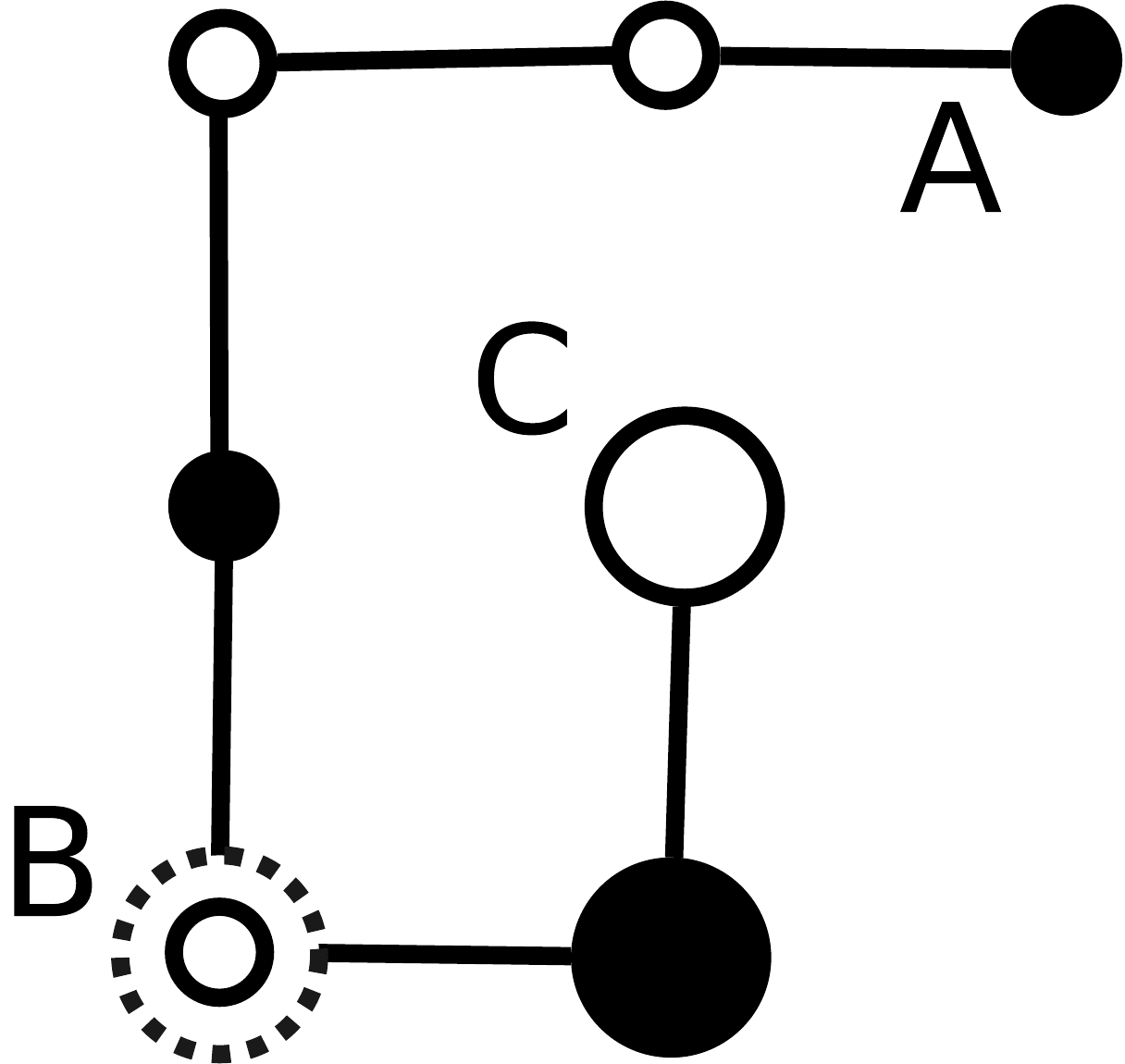}
&\includegraphics[scale=0.12]{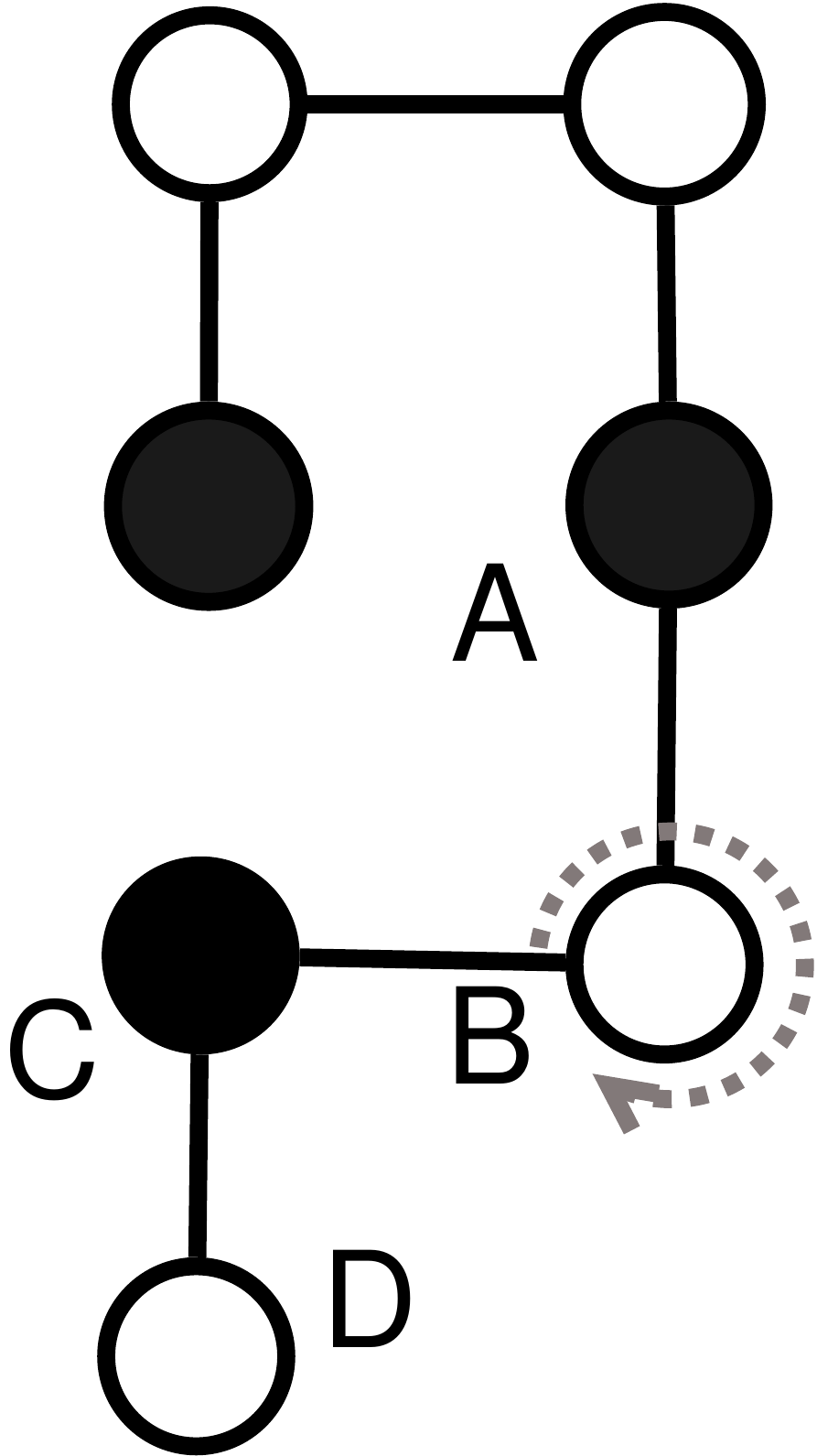}
&\includegraphics[scale=0.12]{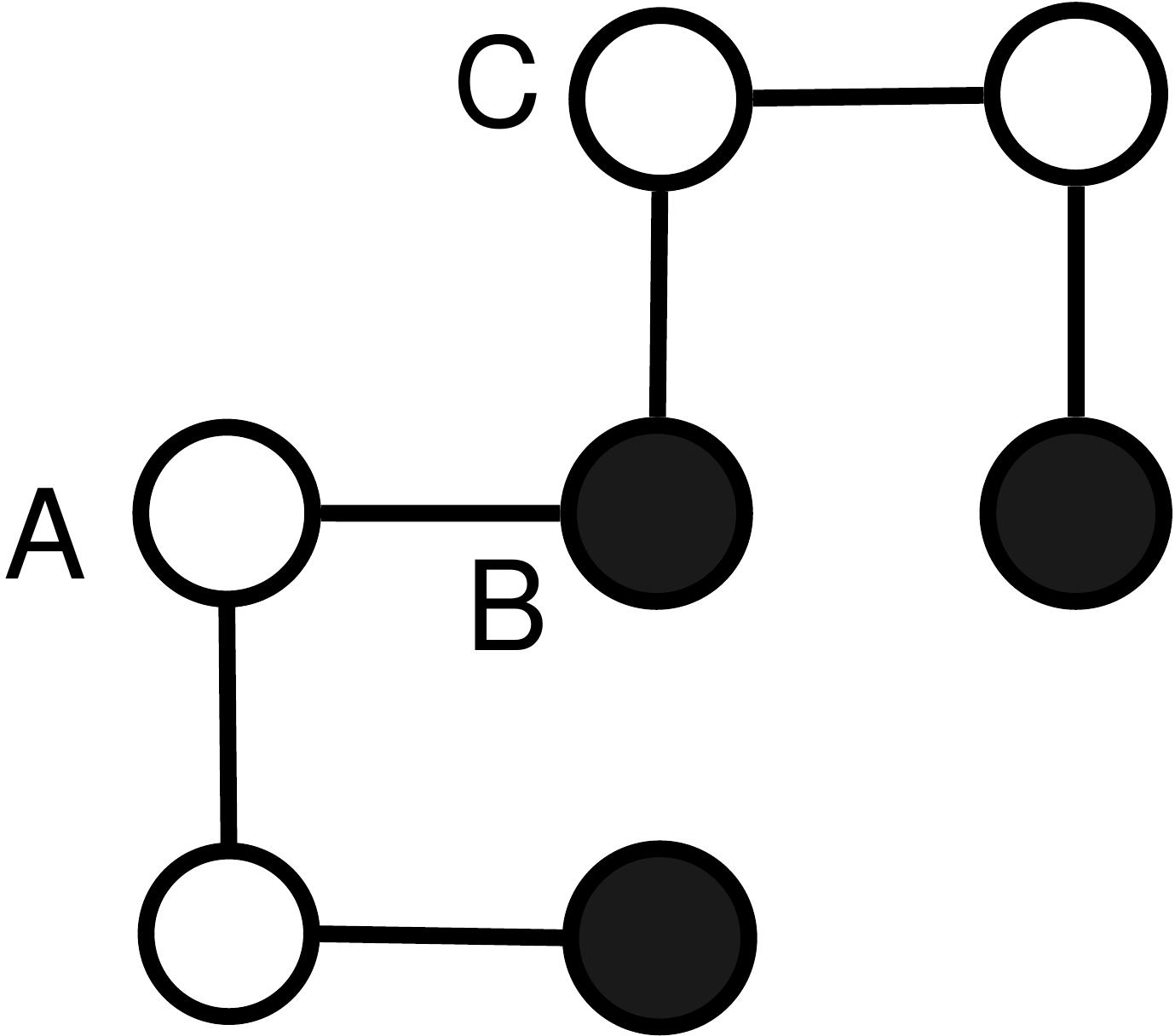}
&\includegraphics[scale=0.12]{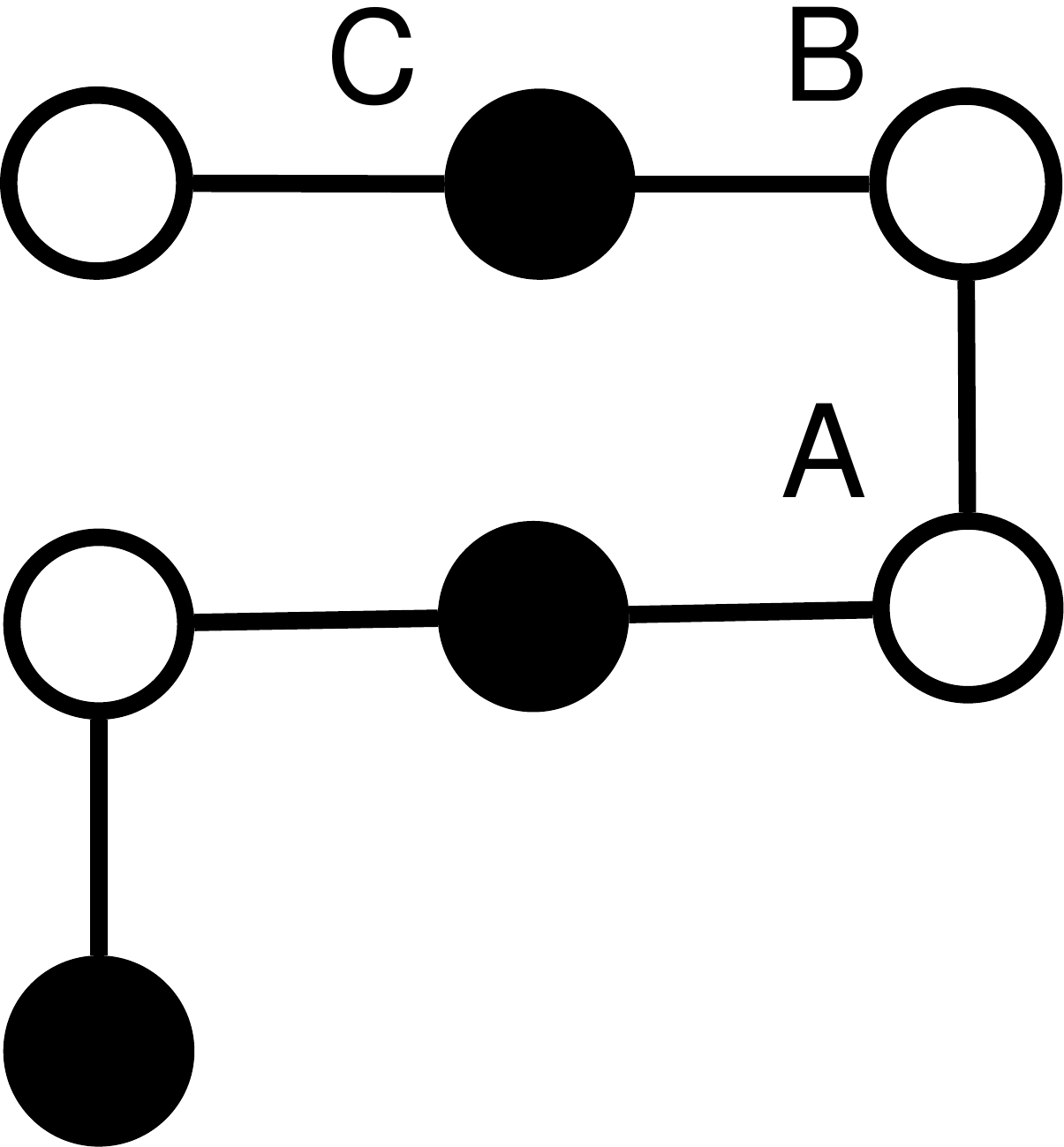}
&\includegraphics[scale=0.12]{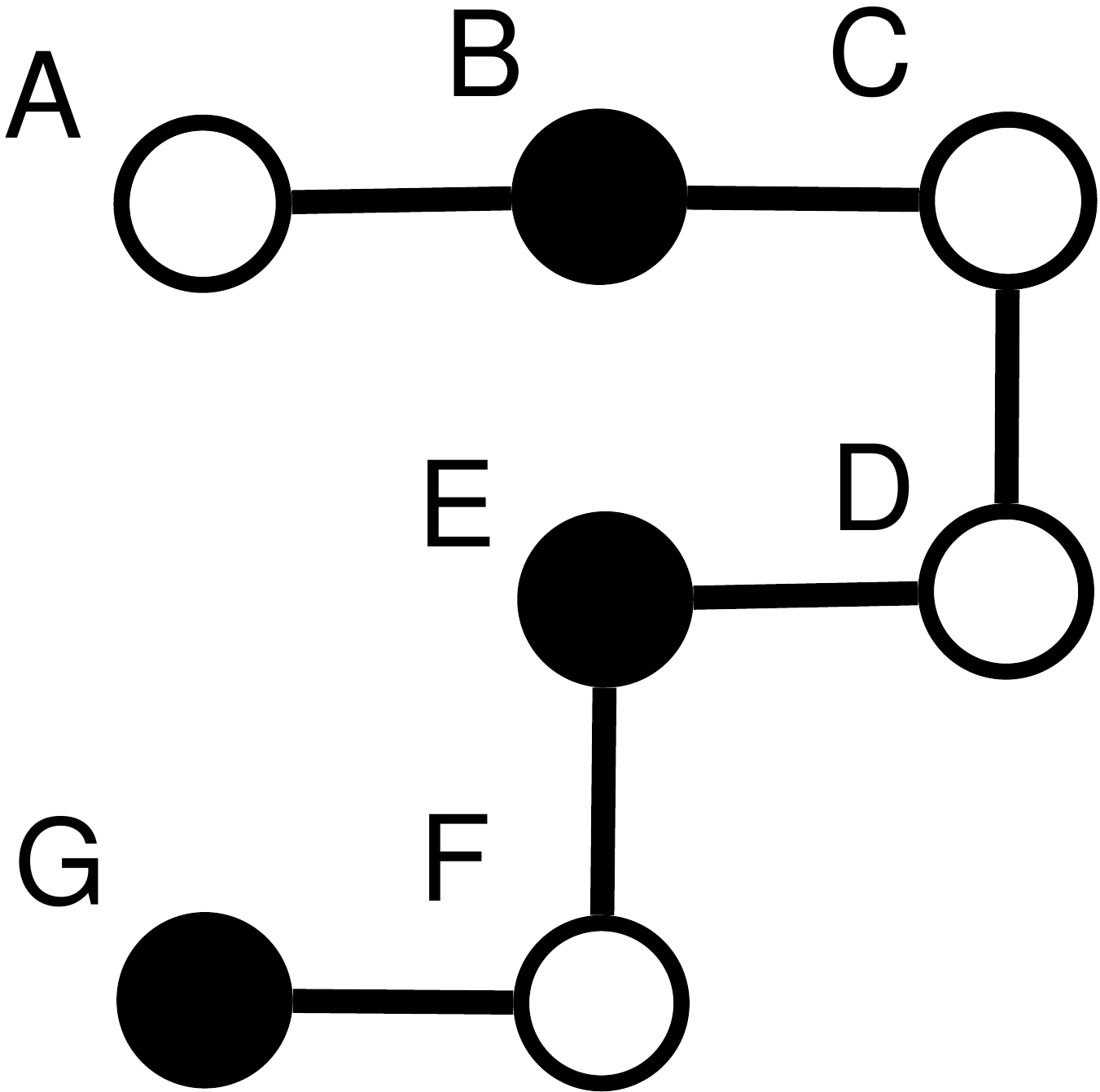}\\
	\multicolumn{3}{c|}{a) single-point crossover~~} & ~~b) rotation~~ & ~~c) diagonal~~ & ~~d) pull~~~~ & ~~~~e) tilt~~ \\
\end{tabular}
\caption{\footnotesize The primitive operators used in our genetic algorithms. The crossover operator applied on two parent conformations to exchange their parts to generate two child conformations (as shown in a) and the mutation operators are applied on single conformation to generate single child conformation (as shown in b, c,d, and e). The operators are implemented in 3D space, however, for simplification and easy understanding the figures are presented in 2D space. The black solid circles represent the hydrophobic {\aas} and others are polar.}
\label{figPrimitiveOp}
\end{figure*}

\subsection{The primitive operators implemented in the GA framework}
The primitive operators that we implemented within the MH\_GA framework are crossover ({\fig}\ref{figPrimitiveOp}a), rotation mutation ({\fig}\ref{figPrimitiveOp}b), diagonal move ({\fig}\ref{figPrimitiveOp}c), pull moves ({\fig}\ref{figPrimitiveOp}d), and tilt moves ({\fig}\ref{figPrimitiveOp}e). The Rotation, diagonal move, pull moves and tilt moves are implemented as mutation operators.

\begin{enumerate}
\item {\bf Crossover:} At a given crossover point (dotted circle in {\fig}\ref{figPrimitiveOp}a), two parent conformations exchange their parts and generate two children. The success rate of crossover operator decreases with the increase of the compactness of the structure.

\item {\bf Rotation:} One part of a given conformation is rotated around a selected point ({\fig}\ref{figPrimitiveOp}b). This move is mostly effective at the beginning of the search.

\item {\bf Diagonal move:} Given three consecutive amino acids at lattice points $A, B$, and $C$, a diagonal move at position $B$ takes the corresponding amino acid diagonally to a free position ({\fig}\ref{figPrimitiveOp}c). The diagonal moves are very effective on FCC  lattice \cite{Cebrian2008TabuFcc_36,Dotu2011OnLattice_40} points.

\item {\bf Pull moves:} The amino acids at points $A$ and $B$ are pulled to the free points ({\fig}\ref{figPrimitiveOp}d) and the connected amino acids are pulled as well to get a valid conformation. The pull moves \cite{Lesh2003MoveSet_73} are local, complete, and reversible. These are very effective especially when the conformation is compact.

\item {\bf Tilt moves:} Two or more consecutive amino acids connected in a straight line are moved by a tilt move to immediately parallel  lattice positions \cite{Hoque2007PhDThesis_28}. The tilt-moves pull the conformation from both sides until a valid conformation is found. In {\fig}\ref{figPrimitiveOp}e, the amino acids at points $C$ and $D$ are moved and subsequently other amino acids from both sides are moved as well.
\end{enumerate}

\subsection{Genetic algorithm framework}
The {\pcode} of MH\_GA framework is presented in {\alg}\ref{algGAFramework}. It uses a set of primitive operators ({\fig}\ref{figPrimitiveOp}) in an exhaustive generation approach to diversify the search, a  hydrophobic core-directed macro-mutation operator to intensify the search, and a random-walk algorithm to recover from the stagnation. Like other search algorithms, it requires initializing the population and the solutions need to be evaluated in each iteration.

\begin{algorithm}[!h]
\footnotesize\sf
	\SetKwData{size}{popSize}
	\SetKwData{op}{op}
	\SetKwData{c}{c}
	\SetKwData{cprime}{c$^\prime$}
	\SetKwData{opR}{opR}
	\SetKwData{curP}{curP}
	\SetKwData{newP}{newP}
	\SetKwData{rwCount}{rwCount}
	\SetKwData{loopCount}{mmCount}
	\SetKwFunction{mutation}{mutation}
	\SetKwFunction{initPopulation}{initialise}
	\SetKwFunction{selectOperator}{selectOperator}
	\SetKwFunction{mutConf}{doMutation}
	\SetKwFunction{macroMut}{doHCDMacroMutation}
	\SetKwFunction{crossConf}{doCrossover}
	\SetKwFunction{randomConfs}{randomConfs}
	\SetKwFunction{full}{full}
	\SetKwFunction{add}{add}
	\SetKwFunction{improved}{improved}
	\SetKwFunction{randomWalk}{goRandomWalk}
	\SetKwFunction{bestConformation}{bestConformation}
	\tcc{{\inParam}Protein/Amino acid sequence, $\size:$  Population size; $\opR$: Operator selection probabilities}
	\tcc{{\outParam}Global best conformation}
	\tcc{{\variables}$\op$: Operators; $\c$, $\cprime$: Conformations; 		
		$\curP$, $\newP$: Current and new populations;
		$\loopCount$: Macro-mutation counter;
		$\rwCount$: Non-improving random-walk counter}
	\vspace{1ex}
	$\curP\leftarrow$\initPopulation
	\label{callInitialise}\;
	\Repeat{(termination criteria)}{
	$\op\leftarrow$, \selectOperator($\opR$)\;

	\uIf{(\op is crossover)}{\tcc{** go for crossover}
		\While{ ($\neg$ \full($\newP$))} {%
			$\c,\cprime\leftarrow$ \randomConfs($\curP$)\;
			\newP.\add(\crossConf)\label{callXOver}\;
		}
	}
	\uElseIf{(\op is mutation)}{\tcc{** go for mutation}
		\ForEach{($\c \in \curP$)}
		{
			\newP.\add(\mutConf)\label{callMutation}\;
		}
	}
	\Else{\tcc{** go for macro-mutation}
		\ForEach{($\c \in \curP$)}
		{
			\newP.\add(\macroMut)\label{callMacroMutation}\;
		}
	}

	\If{ ($\neg$ \improved($\newP$,$\rwCount$))}{
		$\newP \leftarrow $\randomWalk\label{callRandomWalk}\;
	}
	$\curP \leftarrow \newP$\;
}
\KwRet{ \bestConformation($\curP$)}\;
\caption{\sf MH\_GeneticAlgorithm (MH\_GA)}
\label{algGAFramework}
\end{algorithm}

The algorithm initializes ({\alg}\ref{algGAFramework}: Line 7) the current population with randomly generated individuals. At each generation, it selects a genetic operator based on a given probability distribution to use through the generation ({\alg}\ref{algGAFramework}: Line 9). In fact, we select the operators randomly by giving equal opportunities to all operators. The selected  operator is used in an exhaustive manner ({\alg}\ref{algGAFramework}: Line 11-12 or Line 14-16) to obtain all conformations in the new population. We ensure that no duplicate conformation is added to the new population. The {\sf add()} method (Line 12 or 16 in {\alg}\ref{algGAFramework}) takes care of adding the non-duplicate conformations to the new population. For a given number of generations, if the best conformation in the new population is not better than the best in the current population, our algorithm triggers a random-walk technique ({\alg}\ref{algGAFramework}: Line 18) to diversify the new population. Nevertheless, after each generation, the new population becomes the current population ({\alg}\ref{algGAFramework}: Line 19); and the search continues. Finally, the best conformation found so far is returned ({\alg}\ref{algGAFramework}: Line 20). Along with MJ potential matrix, the HP energy model is used during move selection by the macro-mutation operator. The macro-mutation operator is used as other mutation operators ({\fig}\ref{figPrimitiveOp}b-e) in MH\_GA. The details of initialization, evaluation of fitness, exhaustive generation, macro-mutation and stagnation recovery schemes are presented below.

\subsubsection*{Initialization}
Our algorithm starts with a feasible set of conformation known as population. We generate initial conformations following a self-avoiding walk on FCC lattice points. The {\pcode} of the algorithm is presented in  Algorithm~\ref{algInitialise}. It places the first amino acid at $(0,0,0)$. It then randomly selects a basis vector to place the successive amino acid at a neighboring free lattice point. The mapping proceeds until a self-avoiding walk is found for the whole protein sequence.

\begin{algorithm}[!h]
	\footnotesize\sf
	\caption{{\sf initialise}}
	\label{algInitialise}
	\SetKwData{size}{popSize}
	\SetKwData{itj}{j}
	\SetKwData{iti}{i}
	\SetKwData{itp}{p}
	\SetKwData{AA}{AA}
	\SetKwData{conf}{c}
	\SetKwData{fitness}{fitness}
	
	\SetKwData{length}{seqLength}
	\SetKwData{basisVec}{basisVector}
	\SetKwData{initpop}{initPop}
	\SetKwData{node}{point}
	\SetKwFunction{random}{getRandom}

	\SetKwFunction{AminoAcid}{aminoAcid}
	\SetKwFunction{add}{add}
	\SetKwFunction{evaluate}{evaluate}
	
	\tcc{Is called from {\alg}\ref{algGAFramework} in {\lin}\ref{callInitialise}}
	\tcc{{\inParam}Protein/Amino acid sequence, FCC basis vectors, $\size:$ Population size}
	\tcc{{\outParam} Initial population}
	\tcc{{\variables}$\AA:$ Array of amino acid; $\conf:$ Conformations;		
		$\node:$ Unoccupied point on 3D FCC Lattice space}

	\For{\emph{($\itp=1;~\itp\leq\size;~\itp$++)}}{
		$\AA[0]\longleftarrow$ \AminoAcid(0,0,0)\;
		\For{a number of times}{
			\For{(\iti=1 \KwTo \length-1)}{
				$\itj \longleftarrow$ \random(12)\;
				$\node \longleftarrow \AA[\iti-1]+\basisVec[\itj]$\;
				\uIf {\node is not free}{
					break\;
				}\Else{
					$\AA[\iti]\longleftarrow$ \AminoAcid($\node$)\;
				}
			}
		}
		\uIf {full structure found}{
			\conf.AminoAcid$\longleftarrow$\AA[~]\;
		}\Else{
			\conf$\longleftarrow$ a deterministic structure\;
		}
		\conf.\fitness$\longleftarrow$\evaluate(c.\AminoAcid)\label{callEvaluate}\;
		\initpop.\add(\conf)\;
	}
	\KwRet{\initpop}
\end{algorithm}

\subsubsection*{Evaluate the fitness}
For each iteration, the conformation is evaluated by calculating the contacts (topological neighbor) potentials where the two amino acids are non-consecutive. The pseudo-code in Algorithm~\ref{algEvaluation} presents the algorithm for calculating the interaction energy of a given conformation. The contact potentials are found in MJ potential matrix~\cite{Miyazawa1985T20_10} (\emph{see} Table~\ref{mj_matrix}).

\begin{algorithm}[!h]
\footnotesize\sf
	\caption{{\sf evaluate}}
	\label{algEvaluation}
	\SetKwData{itj}{j}
	\SetKwData{iti}{i}
	\SetKwData{pointi}{pointI}
	\SetKwData{pointj}{pointJ}
	\SetKwData{sqrD}{sqrD}
	\SetKwData{AA}{AA}
	\SetKwData{conf}{c}
	\SetKwData{fitness}{fitness}
	\SetKwFunction{getSqrDist}{getSqrDist}
	\SetKwData{length}{seqLength}
	\tcc{Is called from {\alg}\ref{algInitialise} in {\lin}\ref{callEvaluate}}
	\tcc{{\inParam}MJ energy matrix($20\times20$), $\AA:$ Array of amino acid}
	\tcc{{\outParam} Fitness of the structure}
	\tcc{{\variables} $seqLength:$ Sequence length;		
		$\pointi, pointj:$ Occupied point on 3D FCC Lattice space} 		
	$fitness\longleftarrow0$\\
	\For{(\iti=0 \KwTo $\length-1$)}{
		\For{(\itj=\iti+2 \KwTo $\length-1$)}{
			$\pointi\longleftarrow$ \AA[$\iti$]\;
			$\pointj\longleftarrow$ \AA[$\itj$]\;
			$\sqrD \longleftarrow$ \getSqrDist($\pointi,\pointj$)\;
			\If{\sqrD=2}{
				$\fitness \longleftarrow \fitness+${\sf $E_\mathrm{bm}$[$\iti$][$\itj$]}\;
			}
		}
	}
	return $\fitness$\;
\end{algorithm}

\subsubsection*{Exhaustive generation}
Unlike standard genetic algorithm, in MH\_GA, the randomness is reduced significantly by applying exhaustive generation approach.  For mutation operators, MH\_GA adds one resultant conformation to the new population that corresponds to \emph{each} conformation in the current population. Operators are applied to all possible point ({\alg}\ref{algMutation}) exhaustively until finding a better solution than the parent. If no better solution is found, the parent survives through the next generation. On the other hand, for crossover operators, two resultant conformations are added to the new population from two randomly selected parent conformations. Crossover operators generate child conformations by applying the crossover operator in all possible points ({\alg}\ref{algCrossover}) on two randomly selected parents. The best two conformations from the parents and the children are then become the resultant conformations for the next generation.

\begin{algorithm}[!h]
\footnotesize\sf
	\caption{\sf doMutation}
	\label{algMutation}
	
	\SetKwData{pos}{pos}
	\SetKwData{conf}{conf}
	\SetKwData{c}{c}
	\SetKwData{mcnfs}{offspring}
	\SetKwFunction{add}{add}
	\SetKwData{length}{seqLength}
	\SetKwFunction{appOp}{applyOperator}
	\SetKwFunction{bcnf}{bestConformation}
	
	\tcc{Is called from {\alg}\ref{algGAFramework} in {\lin}\ref{callMutation}}
	\tcc{{\inParam}$\conf:$  Conformation}
	\tcc{{\outParam}Best mutated conformation}
	\tcc{{\variables}$\c:$ Conformation; $\mcnfs:$ List of type conformation}

	$\mcnfs.\add(\conf)$\;
	\ForEach{($1 \leq \pos \leq \length$)}{
		$\c \leftarrow \appOp(\conf,\pos)$\;
		$\mcnfs.\add(\c)$\;
	}
	\KwRet{$\bcnf(\mcnfs)$}\;
\end{algorithm}

\begin{algorithm}[!h]
\footnotesize\sf
	\caption{\sf doCrossover}
	\label{algCrossover}
	
	\SetKwData{pos}{pos}
	\SetKwData{conf}{c{$_1$}}
	\SetKwData{confx}{c{$_2$}}
	\SetKwData{c}{c}
	\SetKwData{cprime}{c$^\prime$}
	\SetKwData{ccnfs}{offspring}
	\SetKwFunction{add}{add}
	\SetKwData{seqLength}{seqLength}
	\SetKwFunction{random}{random}
	\SetKwFunction{appOp}{applyOperator}
	\SetKwFunction{bcnf}{best2Conformations}
	
	\tcc{Is called from {\alg}\ref{algGAFramework} in {\lin}\ref{callXOver}}
	\tcc{{\inParam}$\conf$ and $\confx:$  Conformations}
	\tcc{{\outParam}Best two conformations after crossover}
	\tcc{{\variables}$\c,\cprime:$ Conformations; $\ccnfs:$ List of type conformation}
	
	$\ccnfs.\add(\conf,\confx)$\;
	\ForEach{($1 \leq \pos \leq \seqLength$)}{
		$\c,\cprime \leftarrow \appOp(\conf,\confx,\pos)$\;
		$\ccnfs.\add(\c, \cprime)$\;
	}
	\KwRet{$\bcnf(\ccnfs)$}\;
\end{algorithm}

\begin{figure}[!h]
\centering
\includegraphics[width=.75\textwidth]{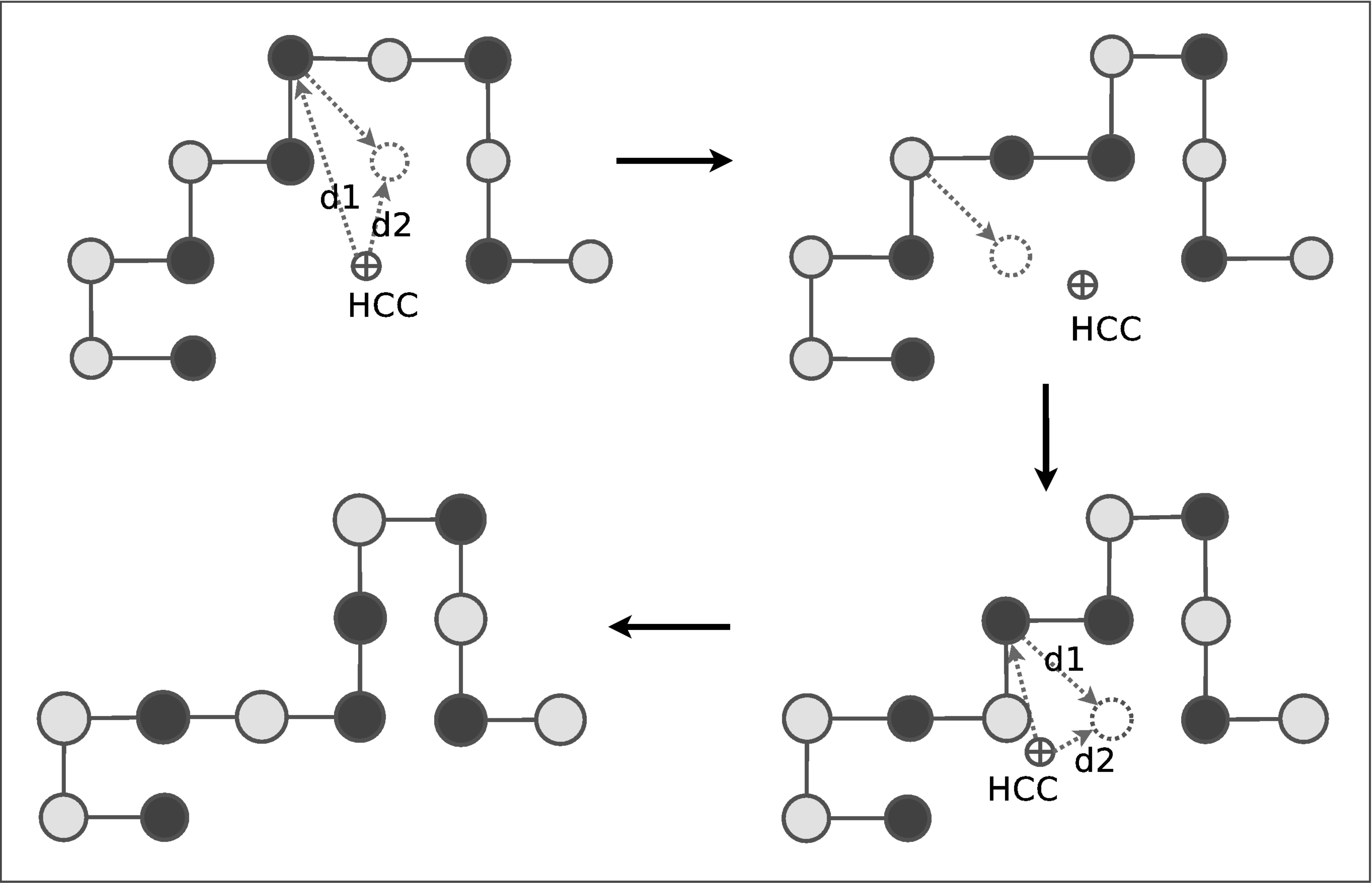}
\caption{\footnotesize A macro-mutation operator repeatedly used diagonal moves. The moves of an amino acid are guided by the distance of current position (d$_1$) and the distance of target position (d$_2$) from the HCC. The operator is implemented in 3D space, however, for simplification and easy understanding, the figures are drawn in 2D space.}
\label{figMacroMut}
\end{figure}

\subsubsection*{Macro-mutation operator}
Protein structures have hydrophobic cores (H-Core) that hide the hydrophobic amino acids from water and expose the polar amino acids to the surface to be in contact with the surrounding water molecules \cite{Yue1993Sequence_74}. H-core formation is an important objective of HP based PSP. Macro-mutation operator is a composite operator ({\fig}\ref{figMacroMut}) that uses a series of diagonal-moves ({\fig}\ref{figPrimitiveOp}c) on a given conformation to build the H-core around the {\hcc} (HCC). The macro-mutation squeezes the conformation and quickly forms the H-core. In MH\_GA, macro-mutation is used as other mutation operators. The {\alg}\ref{algMacroMutation} presents the {\pcode} of macro-mutation algorithm.

\begin{algorithm}[!h]
\footnotesize\sf
	\caption{\sf doHCDMacroMutation}
	\label{algMacroMutation}	
	\SetKwData{hcc}{hcc}
	\SetKwData{dold}{d$_{old}$}
	\SetKwData{dnew}{d$_{new}$}
	\SetKwData{conf}{C}
	\SetKwData{iti}{i}
	\SetKwData{itj}{j}
	\SetKwData{repeat}{repeat}
	\SetKwData{type}{T}
	\SetKwData{point}{point}
	\SetKwData{AA}{AA}
	\SetKwFunction{typ}{typeOf}
	\SetKwFunction{ift}{if}
	\SetKwFunction{els}{else}
	\SetKwFunction{findHCC}{findHCC}
	\SetKwFunction{bern}{bernoulli}
	\SetKwFunction{getDist}{getDistance}
	\SetKwData{len}{seqLength}
	\SetKwFunction{cont}{continue}
	\SetKwFunction{dmove}{applyDiagonalMove($\AA[\itj],\point$)}
	\SetKwFunction{getFreeN}{findFreePoint}
	
	\tcc{Is called from {\alg}\ref{algGAFramework} in {\lin}\ref{callMacroMutation}}
	\tcc{{\inParam}HP energy matrix($2\times2$), \conf:  Conformation; \repeat: Loop counter}
	\tcc{{\outParam} Mutated conformation}
	\tcc{{\variables}\type: Either hydrophobic (H) or polar (P); \AA: Array of amino acid}
	
	\AA[~]$\longleftarrow$\conf.AminoAcid[~]\;
	\For{$\iti=1$ \KwTo repeat}{
		$\type\longleftarrow$ P \ift $\bern(p)$, \els H\\
		$\AA[\itj]:\itj$th amino acid in conformation\\
		$\point$: unoccupied new position for $\AA[\itj]$\\
		\hcc$\longleftarrow$\findHCC()\\
		
		\ForEach{$\itj: \typ(\AA[\itj]) = \type$}
		{
			\dold$\longleftarrow$\getDist($\AA[\itj]$,hcc)\\
			\If{ $\type=$ P}{
				$\point\longleftarrow$\getFreeN($\AA[\itj]$)\\
				\dmove\\
			}
			\Else{
				$\point\longleftarrow$\getFreeN($\AA[\itj]$)\\
				\dnew$\longleftarrow$\getDist($\point$,hcc)\\
				\If{\dnew$\leq$\dold}{
					\dmove\\
					break
				}
			}	
		}
	}
	\conf.AminoAcid[~]$\longleftarrow$\AA[~]\;
	\KwRet{\conf}
	\end{algorithm}
	
In macro-mutation, the HCC is calculated by finding arithmetic means of ${x}$, ${y}$, and ${z}$ coordinates of all H amino acids. In macro-mutation, for a given number of iterations, diagonal moves apply repeatedly either at each P- or at each H-type amino acid positions. Whether to apply the diagonal move on P- or H-type amino acids is determined by using a \emph{Bernoulli} distribution ({\alg}\ref{algMacroMutation} : Line 2) with probability $p$ (intuitively we use $p=20\%$ for P-type amino acids). For a P-type amino acid, the first successful diagonal move is considered. However, for a H-type amino acid, the first successful diagonal move that does not increase the Cartesian distance of the amino acid from the HCC is taken. All the amino acids are traversed and the successful moves are applied as one composite move.

\subsection{Stagnation recovery}
Like other search algorithm, GA can get stuck in the local minima or, can be stalled. Stall condition can occur when similarities with the chromosomes in GA increases heavily and the operators are unable to produce better diverse solutions. Further, with the PSP search, resulting solutions become phenotypically compact which reduce the likelihood of producing better solution from the population due to harder self-avoid-walk (SAW) constraints~\cite{hoque2011twin_17, higgs2012refining_77, hoque2007generalized_78}. It would rather require very intelligent moves to reform into another competitive compact SAW. To deal with such situation, we apply the following two actions:

\subsubsection*{Removing duplicates}
In genetic algorithm it has been observed that with increasing generations, the similarity among the individuals within the population increases. In worst case scenario,  all the individuals become similar and forces the search to stall in the local minima. In our approach, we remove duplicates from each generation to maintain the diversity of the population. During exhaustive generation, we check the existence of the newly generated child in the new population. If it does not exist then the new solution is added to the new population list. Our approach reduces the frequency of stagnations.

\subsubsection*{Applying random-walk}

Sometimes, early convergence leads the search towards the stagnation situation. In the HP energy model, premature H-cores are observed at local minima. To break these H-cores, in MH\_GA ({\alg}\ref{algGAFramework} : Line 18), a random-walk algorithm (\alg\ref{algRandomWalk}) is applied. This algorithm uses pull moves \cite{Lesh2003MoveSet_73} (as shown in Figure \ref{figPrimitiveOp}d) to break the H-core. We use pull-moves because they are complete, local, and reversible. Successful pull moves never generate infeasible conformations. During pulling,  energy level and structural diversification are observed to maintain balance among these two. We allow energy level to change within 5\% to 10\% that changes the structure from 10\%  to 75\% of the original. We try to accept the conformation that is close to the current conformation in terms of the energy level but as far as possible in structural diversity, and which is determined by the function {\sf checkDiversity()} in {\alg}\ref{algRandomWalk} at Line 5. For genetic algorithm, random-walk is very effective \cite{Rashid2012RW_76} to recover from stagnation.

\begin{algorithm}[!h]
	\footnotesize\sf	
	\SetKwData{pos}{i}
	\SetKwData{seqLength}{seqLength}
	\SetKwData{conf}{c}
	\SetKwData{confx}{c$^\prime$}
	\SetKwData{AA}{AA}
	\SetKwData{pct}{pct}
	\SetKwData{inpop}{inPop}
	\SetKwData{outpop}{outPop}
	\SetKwData{isFound}{isFound}
	\SetKwFunction{checkDiversity}{checkDiversity}
	\SetKwFunction{add}{add}
	
	\tcc{Is called from {\alg}\ref{algGAFramework} in {\lin}\ref{callRandomWalk}}
	\tcc{{\inParam}\inpop: Current population; \pct: Changed percentage (\%)}
	\tcc{{\outParam} New diverged population}
	\tcc{{\variables} \outpop: New diverged population; ~~~~~ \AA: Array of amino acid; $\conf,\confx:$ Conformations}
	
	\ForEach{($\conf \in \inpop$)}
		{
			$\isFound\longleftarrow$ false\;
			\AA[~]$\longleftarrow$\conf.AminoAcid[~]\;
			\While{($\neg\isFound$)}{
			\For{\emph{($\pos=1;~\pos\leq\seqLength~\&~\neg\isFound;~\pos$++)}}{
				applyPullMove($\AA[\pos]$)\;
				\confx.AminoAcid[~]$\longleftarrow$\AA[~]\;
				$isFound\longleftarrow$ \checkDiversity(\conf,\confx,\pct)\;
			}
			}
			
			\outpop.\add(\confx)\;
		}	

	\KwRet{\outpop}
	\caption{{\sf goRandomWalk}}
	\label{algRandomWalk}
\end{algorithm}

\begin{figure}[!h]
\centering

\includegraphics[width=15cm]{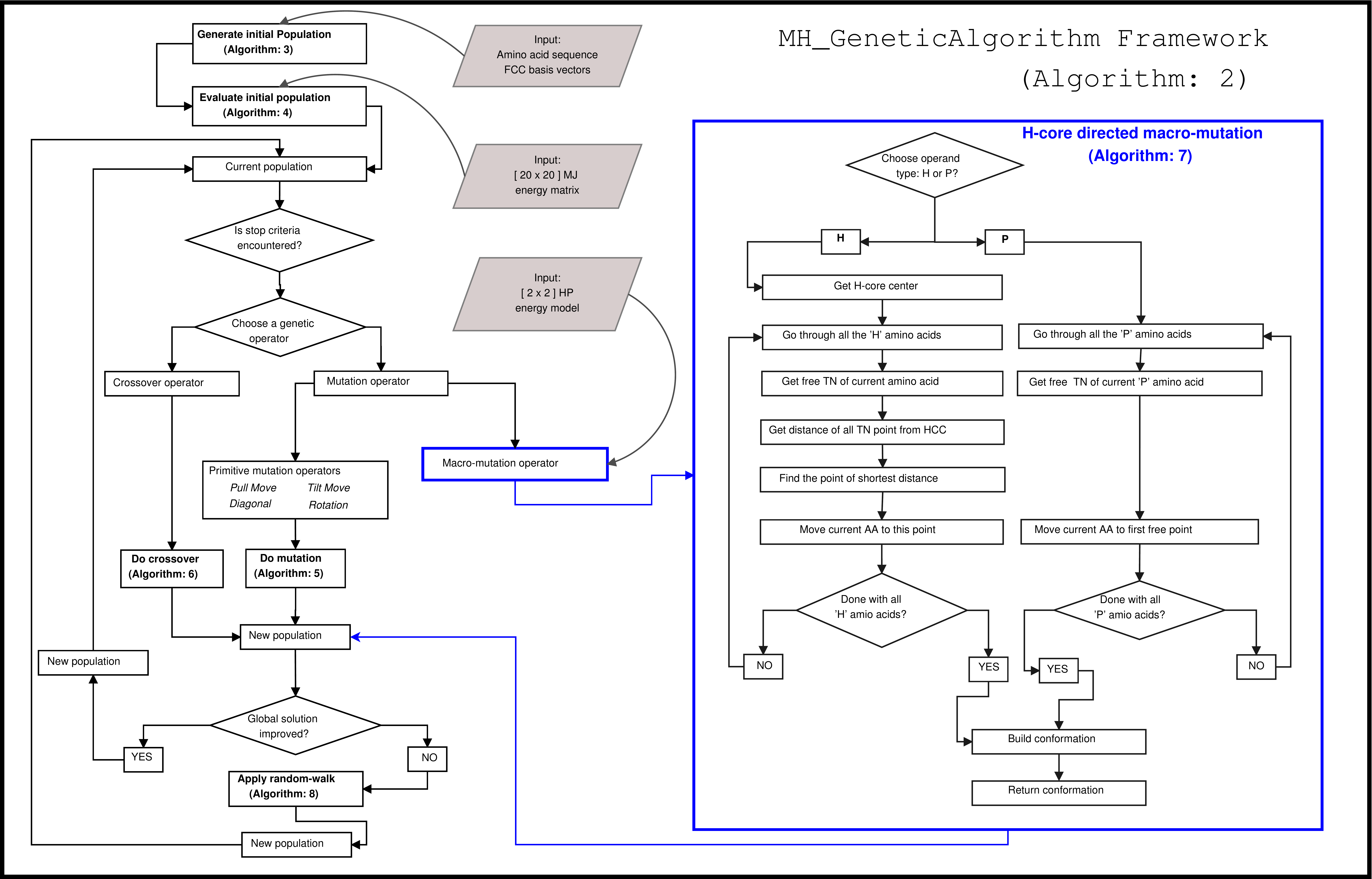}

\caption{\footnotesize A complete overview of our algorithmic approach. The macro-mutation procedure is described step by step (inside the blue box). The procedural sub blocks are marked in bold along with the corresponding labels of the algorithms described above.}
\label{figFlowChart}
\end{figure}

The complete flow of MH\_GeneticAlgorithm (Algorithm:~\ref{algGAFramework}) is graphically presented in Figure~\ref{figFlowChart}. Further, it describes the steps taken within macro mutation procedure (Algorithm:~\ref{algMacroMutation}).

\section{Performance Evaluation}
\label{secPerfEvaluation}
To compare and evaluate the performance of the proposed PSP predictor with respect to the state-of-the-art approaches, we used the measures \textit{Relative Improvement (RI)} and \textit{RMSD} comparisons. They are defined below:

\subsubsection*{Relative Improvement (RI)}
The difficulty to improve energy level is increased as the predicted energy level approaches to a known lower bound of a given protein. For example, if the lower bound of free energy of a protein is $-100$, the efforts to improve energy level from $-80$ to $-85$ is much less than that to improve energy level from $-95$ to $-100$ though the change in energy is the same ($-5$). The RI computes the relative improvements that our algorithm (target,\emph{ t}) achieved w.r.t. the state-of-the-art approaches (reference\emph{, r}).

For each protein, the relative improvement of the target ($t$) w.r.t. the reference ($r$) is calculated using the formula in Equation~\ref{eqImp}, where $E_t$ and $E_r$ denote the average energy values achieved by target and reference respectively.

\begin{footnotesize}
\begin{equation}
 	{\sf RI} = \frac{E_t - E_r}{E_r} \times 100\%
	\label{eqImp}
\end{equation}
\end{footnotesize}

\subsubsection*{RMSD comparison}
The root mean square deviation (RMSD) is frequently used to measure the differences between values predicted by a model and the values actually observed. We compare the predicted structures obtained by our approach with the state-of-the-art approaches by measuring the root-mean-square w.r.t. the native structures from PDB. For any given structure the root-mean-square is calculated using Equation~\ref{equRMSD},

\begin{equation}
	\mathrm{RMSD}= \sqrt\frac{\sum_{i=1}^{n-1}\sum_{j=i+1}^{n} (d_{ij}^{\mathrm{p}}-d_{ij}^{\mathrm{n}})^2}{n\times(n-1)/2}
	\label{equRMSD}
\end{equation}

where $d_{ij}^\mathrm{p}$ and $d_{ij}^\mathrm{n}$ denote the distances between $i$th and $j$th amino acids respectively in the predicted structure and the native structure of the protein. The average distance between two $\alpha$-Carbons in native structure is $3.8$\AA. To calculate root-mean-square, the distance between two neighbor lattice points is considered as $3.8$\AA.

\begin{table}[!h]
\centering
\begin{scriptsize}
	\caption{\footnotesize The benchmark proteins used in our experiments.}
	\label{tabBenchmark}
	\sf
	\setlength{\tabcolsep}{5pt}
	\rowSpace
\begin{tabular}{ccp{10cm}}
		\hline
		{\bf ID}&{\bf Len}&{\bf Protein sequence}\\
		\hline
		{\it 4RXN}&{54}&\texttt{MKKYTCTVCGYIYNPEDGDPDNGVNPGTDFKDIPDDWVCPLCGVGKDQFEEVEE}\\
		\hline
		{\it 1ENH}&{54}&\texttt{RPRTAFSSEQLARLKREFNENRYLTERRRQQLSSELGLNEAQIKIWFQNKRAKI}\\
		\hline
		{\it 4PTI}&{58}&\texttt{RPDFCLEPPYTGPCKARIIRYFYNAKAGLCQTFVYGGCRAKRNNFKSAEDCMRTCGGA}\\
		\hline
		{\it 2IGD}&{61}&\texttt{MTPAVTTYKLVINGKTLKGETTTKAVDAETAEKAFKQYANDNGVDGVWTYDDATKTFTVTE}\\
		\hline
		{\it 1YPA}&{64}&\texttt{MKTEWPELVGKAVAAAKKVILQDKPEAQIIVLPVGTIVTMEYRIDRVRLFVDKLDNIAQVPRVG}\\
		\hline
		{\it 1R69}&{69}&\texttt{SISSRVKSKRIQLGLNQAELAQKVGTTQQSIEQLENGKTKRPRFLPELASALGVSVDWLLNGTSDSNVR}\\
		\hline
		{\it 1CTF}&{74}&\texttt{AAEEKTEFDVILKAAGANKVAVIKAVRGATGLGLKEAKDLVESAPAALKEGVSKDDAEALKKALEEAGAEVEVK}\\
		\hline
		{\it 3MX7}&{90}&\texttt{MTDLVAVWDVALSDGVHKIEFEHGTTSGKRVVYVDGKEEIRKEWMFKLVGKETFYVGAAKTKATINIDAISGFA YEYTLEINGKSLKKYM}\\
		\hline
		{\it 3NBM}&{108}&\texttt{SNASKELKVLVLCAGSGTSAQLANAINEGANLTEVRVIANSGAYGAHYDIMGVYDLIILAPQVRSYYREMKVDA ERLGIQIVATRGMEYIHLTKSPSKALQFVLEHYQ}\\
		\hline
		{\it 3MQO}&{120}&\texttt{PAIDYKTAFHLAPIGLVLSRDRVIEDCNDELAAIFRCARADLIGRSFEVLYPSSDEFERIGERISPVMIAHGSY ADDRIMKRAGGELFWCHVTGRALDRTAPLAAGVWTFEDLSATRRVA}\\
		\hline
		{\it 3MRO}&{142}&\texttt{SNALSASEERFQLAVSGASAGLWDWNPKTGAMYLSPHFKKIMGYEDHELPDEITGHRESIHPDDRARVLAALKA
		HLEHRDTYDVEYRVRTRSGDFRWIQSRGQALWNSAGEPYRMVGWIMDVTDRKRDEDALRVSREELRRL}\\
		\hline
		{\it 3PNX}&{160}&\texttt{GMENKKMNLLLFSGDYDKALASLIIANAAREMEIEVTIFCAFWGLLLLRDPEKASQEDKSLYEQAFSSLTPREA EELPLSKMNLGGIGKKMLLEMMKEEKAPKLSDLLSGARKKEVKFYACQLSVEIMGFKKEELFPEVQIMDVKEYL KNALESDLQLFI}\\
		\hline
		{\it 2J6A}&{135}&\texttt{MKFLTTNFLKCSVKACDTSNDNFPLQYDGSKCQLVQDESIEFNPEFLLNIVDRVDWPAVLTVAAELGNNALPPT KPSFPSSIQELTDDDMAILNDLHTLLLQTSIAEGEMKCRNCGHIYYIKNGIPNLLLPPHLV}\\
		\hline
		{\it 2HFQ}&{85}&\texttt{MQIHVYDTYVKAKDGHVMHFDVFTDVRDDKKAIEFAKQWLSSIGEEGATVTSEECRFCHSQKAPDEVIEAIKQN GYFIYKMEGCN}\\
		\hline
		{\it 3MSE}&{180}&\texttt{GISPNVLNNMKSYMKHSNIRNIIINIMAHELSVINNHIKYINELFYKLDTNHNGSLSHREIYTVLASVGIKKWD INRILQALDINDRGNITYTEFMAGCYRWKNIESTFLKAAFNKIDKDEDGYISKSDIVSLVHDKVLDNNDIDNFF LSVHSIKKGIPREHIINKISFQEFKDYMLSTF}\\
		\hline
		{\it 3MR7}&{189}&\texttt{SNAERRLCAILAADMAGYSRLMERNETDVLNRQKLYRRELIDPAIAQAGGQIVKTTGDGMLARFDTAQAALRCA LEIQQAMQQREEDTPRKERIQYRIGINIGDIVLEDGDIFGDAVNVAARLEAISEPGAICVSDIVHQITQDRVSE PFTDLGLQKVKNITRPIRVWQWVPDADRDQSHDPQPSHVQH}\\
		\hline
		{\it 3MQZ}&{215}&\texttt{SNAMSVQTIERLQDYLLPEWVSIFDIADFSGRMLRIRGDIRPALLRLASRLAELLNESPGPRPWYPHVASHMRRR VNPPPETWLALGPEKRGYKSYAHSGVFIGGRGLSVRFILKDEAIEERKNLGRWMSRSGPAFEQWKKKVGDLRDFG PVHDDPMADPPKVEWDPRVFGERLGSLKSASLDIGFRVTFDTSLAGIVKTIRTFDLLYAEAEKGS}\\		
		\hline
		{\it 3NO3}&{238}&\texttt{GKDNTKVIAHRGYWKTEGSAQNSIRSLERASEIGAYGSEFDVHLTADNVLVVYHDNDIQGKHIQSCTYDELKDLQ LSNGEKLPTLEQYLKRAKKLKNIRLIFELKSHDTPERNRDAARLSVQMVKRMKLAKRTDYISFNMDACKEFIRLC PKSEVSYLNGELSPMELKELGFTGLDYHYKVLQSHPDWVKDCKVLGMTSNVWTVDDPKLMEEMIDMGVDFITTDL PEETQKILHSRAQ}\\
		\hline
		{\it 3NO7}&{248}&\texttt{MGSDKIHHHHHHENLYFQGMTFSKELREASRPIIDDIYNDGFIQDLLAGKLSNQAVRQYLRADASYLKEFTNIYA MLIPKMSSMEDVKFLVEQIEFMLEGEVEAHEVLADFINEPYEEIVKEKVWPPSGDHYIKHMYFNAFARENAAFTI
		AAMAPCPYVYAVIGKRAMEDPKLNKESVTSKWFQFYSTEMDELVDVFDQLMDRLTKHCSETEKKEIKENFLQSTI HERHFFNMAYINEKWEYGGNNNE}\\
		\hline
		{\it 3ON7}&{280}&\texttt{GMKLETIDYRAADSAKRFVESLRETGFGVLSNHPIDKELVERIYTEWQAFFNSEAKNEFMFNRETHDGFFPASIS ETAKGHTVKDIKEYYHVYPWGRIPDSLRANILAYYEKANTLASELLEWIETYSPDEIKAKFSIPLPEMIANSHKT
		LLRILHYPPMTGDEEMGAIRAAAHEDINLITVLPTANEPGLQVKAKDGSWLDVPSDFGNIIINIGDMLQEASDGY
		FPSTSHRVINPEGTDKTKSRISLPLFLHPHPSVVLSERYTADSYLMERLRELGVL}\\
		\hline
	\end{tabular}	

\end{scriptsize}
\end{table}

\section{Results and discussion}
\label{secResults}
In this section, we discuss the obtained results along with the comparison of the performance of MH\_GeneticAlgorithm with the other state-of-the-art results~\cite{Torres2007GAT20_59, Ullah2010Hybrid_60, Shatabda2013Heuristic_63}. Further, we present an analysis of the results.

\subsection{Benchmark}
In our experiment, the protein instances  are taken from the literatures. The first seven proteins ({\ba},{\bb}, {\bc}, {\bd}, {\be}, {\bfx}, and {\bg}) in Table~\ref{tabBenchmark} are taken from \cite{Ullah2010Hybrid_60} and \cite{Shatabda2013Heuristic_63}, and the next five proteins ({\ca}, {\cb}, {\cc}, {\cd}, and {\ce}) are taken from \cite{Shatabda2013Heuristic_63}.  The two other protein instances in {\tab}\ref{tabGECCO} ({\da} and {\db}) are taken from \cite{Torres2007GAT20_59}.

\begin{table*}[!h]
\caption{\footnotesize The energy values are obtained from different algorithms for the specified energy models. The average values are calculated over $50$ different runs. The bold-faced values indicate the winner (the lower the better).}
\renewcommand{\arraystretch}{1.5}
\centering
\begin{scriptsize}

\setlength{\tabcolsep}{6pt}
\begin{tabular}{cccrrcrrcrrcc}
		\hline
		\multicolumn{3}{c}{\bf }&\multicolumn{6}{c}{\bf The \sota}&\multicolumn{4}{c}{\bf Our approach}\\	
		\cline{4-13}
		\multicolumn{3}{c}{\bf Protein}&\multicolumn{3}{c}{\bf Hybrid \cite{Ullah2010Hybrid_60}}
			&\multicolumn{3}{c}{\bf Local Search \cite{Shatabda2013Heuristic_63}}	
			&\multicolumn{4}{c}{\bf The MH\_GA}\\		
		\cline{4-13}
		\multicolumn{3}{c}{\bf details }& \multicolumn{2}{c}{\bf MJ energy }&{\bf Time}
			& \multicolumn{2}{c}{\bf MJ energy }&{\bf Time}& \multicolumn{2}{c}{\bf MJ energy }&{\bf Time}&{\bf RI}\\		
		\hline
		{\it Seq}&{\it Size}&{\it H}&{\it Best}&{\it Avg}&{\it Avg}&{\it Best}&{\it Avg (r)}&{\it Avg}
			&{\it Best}&{\it Avg (t)}&{\it Avg}&{over \cite{Shatabda2013Heuristic_63}}\\
		\hline	
	{4RXN}&{54}&{27}&{-32.61}&{-30.94}&{1:02:12}&{-33.33}&{-31.21}&{}&{\bf -36.36}&{\bf -33.60}&{}&{7.66\%}\\
	{1ENH}&{54}&{19}&{-35.81}&{-35.07}&{1:02:03}&{-29.03}&{-28.18}&{}&{\bf -38.39}&{\bf -35.67}&{}&{26.58\%}\\
	{4PTI}&{58}&{32}&{-32.07}&{-29.37}&{1:01:26}&{-31.16}&{-28.33}&{}&{\bf -35.65}&{\bf -31.01}&{}&{9.46\%}\\
	{2IGD}&{61}&{25}&{\bf -38.64}&{-32.54}&{1:43:08}&{-32.36}&{-28.29}&{1:00:00}&{-36.49}&{\bf -33.75}&{1:00:00}&{19.30\%}\\
	{1YPA}&{64}&{38}&{{\it n/a}}&{{\it n/a}}&{}&{-33.33}&{-32.15}&{}&{\bf -40.14}&{\bf -36.33}&{}&{13.00\%}\\
	{1R69}&{69}&{30}&{-34.2}&{-31.85}&{1:07:32}&{-33.35}&{-32.20}&{}&{\bf -40.85}&{\bf -36.28}&{}&{12.67\%}\\
	{1CTF}&{74}&{42}&{-38}&{-35.28}&{1:37:44}&{-45.83}&{-40.94}&{}&{\bf -51.5}&{\bf -47.29}&{}&{15.51\%}\\
	\hline
	{3MX7}&{90}&{44}&{{\it n/a}}&{{\it n/a}}&{}&{-44.81}&{-42.32}&{}&{\bf -56.32}&{\bf -50.95}&{}&{20.39\%}\\
	{3NBM}&{108}&{56}&{{\it n/a}}&{{\it n/a}}&{}&{-52.44}&{-49.51}&{}&{\bf -53.66}&{\bf -49.9}&{}&{0.79\%}\\
	{3MQO}&{120}&{68}&{{\it n/a}}&{{\it n/a}}&{}&{\bf -64.04}&{\bf -58.84}&{1:00:00}&{-62.25}&{-54.56}&{1:00:00}&{\it no RI}\\
	{3MRO}&{142}&{63}&{{\it n/a}}&{{\it n/a}}&{}&{-87.38}&{-82.24}&{}&{\bf -90.05}&{\bf -82.32}&{}&{0.10\%}\\
	{3PNX}&{160}&{84}&{{\it n/a}}&{{\it n/a}}&{}&{\bf -103.04}&{\bf -96.86}&{}&{-102.55}&{-88.06}&{}&{\it no RI}\\
	\hline
	{3MSE}&{180}&{83}&{{\it n/a}}&{{\it n/a}}&{}&{{\it n/a}}&{{\it n/a}}&{}&{-92.61}&{-84.60}&{}&{{\it n/a}}\\
	{3MR7}&{189}&{88}&{{\it n/a}}&{{\it n/a}}&{}&{{\it n/a}}&{{\it n/a}}&{}&{-93.65}&{-83.93}&{}&{{\it n/a}}\\
	{3MQZ}&{215}&{115}&{{\it n/a}}&{{\it n/a}}&{}&{{\it n/a}}&{{\it n/a}}&{}&{-104.29}&{-95.22}&{2:00:00}&{{\it n/a}}\\
	{3NO3}&{238}&{102}&{{\it n/a}}&{{\it n/a}}&{}&{{\it n/a}}&{{\it n/a}}&{}&{-122.97}&{-108.70}&{}&{{\it n/a}}\\
	{3NO6}&{248}&{112}&{{\it n/a}}&{{\it n/a}}&{}&{{\it n/a}}&{{\it n/a}}&{}&{-133.95}&{-117.11}&{}&{{\it n/a}}\\
	{3ON7}&{280}&{135}&{{\it n/a}}&{{\it n/a}}&{}&{{\it n/a}}&{{\it n/a}}&{}&{-116.88}&{-96.64}&{}&-{{\it n/a}}\\
	\hline
	\multicolumn{13}{l}{\footnotesize{\bf {\it n/a}} denotes the experimental results are not available.}
\end{tabular}	
\end{scriptsize}
\label{tabSOTAEnergy}
\end{table*}

\begin{table*}[!h]
\caption{\footnotesize The average energy and average RMSD values achieved from two different variants of GA.  The average values are calculated over $50$ different runs. The {bold-faced} values indicate the winner (the lower the better).}
\renewcommand{\arraystretch}{1.5}
\centering
\begin{scriptsize}
	\setlength{\tabcolsep}{7pt}
	\begin{tabular}{ccccccccccccc}
		\hline		
		\multicolumn{3}{c}{\bf }&\multicolumn{4}{c}{\bf The {\sota} GA  \cite{Torres2007GAT20_59}}
			&\multicolumn{6}{c}{\textbf{The MH\_GA}}\\
		\cline{4-13}
		
		\multicolumn{3}{c}{\bf Protein details}&\multicolumn{4}{c}{Reported values}
				&\multicolumn{4}{c}{Average values}&{}&{}\\
		\cline{4-11}
		
		\multicolumn{3}{c}{\bf }&\multicolumn{4}{c}{\bf MJ model}
			&\multicolumn{2}{c}{\bf MJ model}&\multicolumn{2}{c}{\bf MH model}&{}&{\it Gen }\\
		\hline
		{\it Seq}&{\it Size}&{\it H}&{\it Energy }&{\it RMSD}&{\it Pop}&{\it Gen}
		&{\it Energy }&{\it RMSD}&{\it Energy }&{\it RMSD}&{\it Pop}&{\it ( $\leq$)}\\
		\hline	
		
		{2J6A}&{135}&{71}&{-815.82*} &{16.75} &{50}&{20000} &{-59.72}&{9.53}&{\bf -61.40}&{\bf 9.48}&{50}&{\bf 2500}\\
		{2HFQ}&{85}&{38} &{-543.17*} &{12.24} &{50} &{20000}&{-52.13}&{7.48}&{\bf -52.72}&{\bf 7.31}&{50}&{\bf 7000}\\
		\hline
		\multicolumn{13}{l}{\footnotesize{\bf *} the unusual values for MJ energy model.}
	\end{tabular}	
\end{scriptsize}

\label{tabGECCO}
\end{table*}

\subsection{Comparing with the {\sota}}
In the literature we found very few works \cite{Kapsokalivas2009Population_58, Torres2007GAT20_59} that used {\ttx} MJ potential-matrix \cite{Miyazawa1985T20_10} for protein structure prediction on 3D FCC lattice. However, Torres {\etal} \cite{Torres2007GAT20_59} used 3D HCP lattice and Kapsokalivas {\etal} \cite{Kapsokalivas2009Population_58} used 3D cubic lattice in their works for protein mapping. In other works, Ullah {\etal} \cite{Ullah2010Hybrid_60} and Shatabda {\etal} \cite{Shatabda2013Heuristic_63} used 3D FCC lattice with {\ttx} empirical energy matrix by Berrera {\etal} \cite{Berrera2003T20_11}. In fact, we do not have any {\sota} results available for similar model to compare free energy level in a straight way. Therefore, we ran the algorithms used in \cite{Ullah2010Hybrid_60} and \cite{Shatabda2013Heuristic_63} using the MJ energy model \cite{Miyazawa1985T20_10} to compare our results. However, the constraint programming based hybrid approach \cite{Ullah2010Hybrid_60} failed to get any solution for most of the large-sized proteins. In such cases, in {\tab}\ref{tabSOTAEnergy}, the results are denoted by {\it n/a}.

\begin{figure}[!h]
\centering
	\includegraphics[width=8cm,height=4cm]{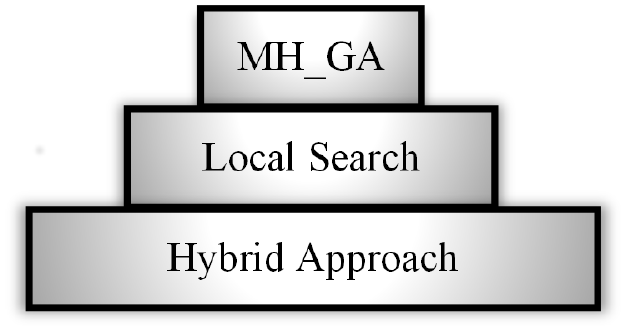}
\vspace{-2ex}
\caption{\footnotesize  The performance hierarchy among the {\sota} approaches and our MH\_GA. Our GA outperforms the other to approaches in \cite{Ullah2010Hybrid_60} and \cite{Shatabda2013Heuristic_63}. }
\label{figPerfHierarchy}
\end{figure}

In {\tab}\ref{tabSOTAEnergy}, we present interaction energy values in two different formats: the global lowest interaction energy (Column {\it Best}) and the average (Column {\it Avg}) of the lowest interaction energies obtained from $50$ different runs. In case of the global best energy, our approach outperforms the {\sota} approaches in \cite{Ullah2010Hybrid_60, Shatabda2013Heuristic_63} on $9$ out of $12$ benchmark proteins. However, in case of average energy, our approach outperforms both of the approaches on $10$ out of $12$ benchmark proteins. Based on the experimental results, the performance hierarchy of the approaches used to validate our MH\_GA is shown in {\fig}\ref{figPerfHierarchy}.

\subsubsection*{Outcome based on Relative Improvement (RI)}
From the Column RI of Table~\ref{tabSOTAEnergy}, we see that for 2 proteins our GA fail to improve over the {\sota}. However, for other 10 proteins it improves the average interaction energy level ranging from $0.10$\% to $26.58$\% for different proteins.

Further, in {\tab}\ref{tabGECCO}, we present another two benchmark proteins taken from a GA based approach \cite{Torres2007GAT20_59}. From the authors of \cite{Torres2007GAT20_59}, we tried to get their implemented codes so that we can run that by ourselves. However, we failed to receive any response from the authors. Therefore, we present the reported values. For fair comparison, we compare the results by generation-wise instead of by running-time.

\begin{table}[!h]
\caption{\footnotesize The best RMSD values reported, are the best amongst the $50$ different runs. The { bold-faced} values indicate the winner (the lower the better).}
\renewcommand{\arraystretch}{1.5}
\centering
\begin{scriptsize}
	\setlength{\tabcolsep}{10pt}
	\rowSpace
		\begin{tabular}{ccccccc}
		\hline
		\multicolumn{3}{c}{\bf Protein details}&{\bf Local Search \cite{Shatabda2013Heuristic_63}}&\multicolumn{3}{c}{\textbf{The MH\_GA}}\\
		\hline
		{\it Seq}&{\it Size}&{\it H}&{\it MJ guided}&{\it HP guided}&{\it MJ guided}&{\it MH guided}\\
		\hline
{4RXN}&{54}&{27}&{5.74}&{\bf 4.70}&{4.83}&{4.76}\\
{1ENH}&{54}&{19}&{5.94}&{\bf 4.42}&{4.75}&{4.81}\\
{4PTI}&{58}&{32}&{\bf 6.02}&{6.18}&{6.24}&{6.06}\\
{2IGD}&{61}&{25}&{7.38}&{7.64}&{6.63}&{\bf 6.53}\\
{1YPA}&{64}&{38}&{6.54}&{\bf 5.17}&{5.52}&{5.39}\\
{1R69}&{69}&{30}&{6.12}&{\bf 4.44}&{4.76}&{4.64}\\
{1CTF}&{74}&{42}&{6.08}&{4.72}&{4.26}&{\bf 4.08}\\
\hline
{3MX7}&{90}&{44}&{8.17}&{\bf 7.10}&{7.21}&{7.20}\\
{3NBM}&{108}&{56}&{6.38}&{5.89}&{5.64}&{\bf 5.37}\\
{3MQO}&{120}&{68}&{6.92}&{6.44}&{6.33}&{\bf 6.38}\\
{3MRO}&{142}&{63}&{8.76}&{7.76}&{7.93}&{\bf 7.64}\\
{3PNX}&{160}&{84}&{8.78}&{7.90}&{8.04}&{\bf 7.60}\\
\hline
{3MSE}&{180}&{83}&{\it n/a}&{20.24}&{\bf 16.05}&{16.98}\\
{3MR7}&{189}&{88}&{\it n/a}&{10.43}&{9.42}&{\bf 9.36}\\
{3MQZ}&{215}&{115}&{\it n/a}&{11.21}&{\bf 8.88}&{9.04}\\
{3NO3}&{238}&{102}&{\it n/a}&{14.49}&{\bf 11.22}&{11.70}\\
{3NO6}&{248}&{112}&{\it n/a}&{13.20}&{13.88}&{\bf 12.04}\\
{3ON7}&{280}&{135}&{\it n/a}&{13.19}&{11.84}&{\bf 11.77}\\
\hline
\end{tabular}	
\end{scriptsize}
\label{tabSOTARMSD}
\end{table}

\subsubsection*{Outcome based on RMSD comparison}
We calculate RMSD of a structure that corresponds to the lowest MJ interaction energy for a particular run. The reported RMSD values in {\tab}\ref{tabSOTARMSD} are the global minimum of 50 runs. In {\tab}\ref{tabGECCO} and {\tab}\ref{tabSOTARMSD}, the {bold-faced} RMSD values indicate the winners for the corresponding proteins.

In {\tab}\ref{tabMinRMSD}, we present corresponding MJ energy values for global minimum RMSD and corresponding RMSD values for global minimum MJ energy values over $50$ runs for each proteins on identical settings. The experimental results show that the global minimum energy in our experiment does not produce minimum RMSD value.

\begin{table*}[!h]
\caption{\footnotesize Corresponding MJ energies for global minimum RMSD and corresponding RMSDs for global minimum MJ energies over $50$ runs for each proteins}
\renewcommand{\arraystretch}{1.5}
\centering
\begin{scriptsize}
	\setlength{\tabcolsep}{3pt}
		\begin{tabular}{cccrrrrrrrrrrrr}
		\hline
		\multicolumn{3}{c}{\bf Protein}&\multicolumn{6}{c}{\small Energy corresponds to RMSD}
			&\multicolumn{6}{c}{\small RMSD corresponds to energy}\\
		\cline{4-15}
		\multicolumn{3}{c}{\bf details}&\multicolumn{2}{c}{\bf HP}
		&\multicolumn{2}{c}{\bf MJ}&\multicolumn{2}{c}{ \bf MH}
		&\multicolumn{2}{c}{\bf HP}
		&\multicolumn{2}{c}{\bf MJ}&\multicolumn{2}{c}{ \bf MH}\\
\hline
{\it Seq}&{\it Size}&{\it H}&{\it rmsd}&{\it En}&{\it rmsd}&{\it En}&{\it rmsd}&{\it En}
	&{\it En}&{\it rmsd}&{\it En}&{\it rmsd}&{\it En}&{\it rmsd}\\
\hline
{4RXN}&{54}&{27}&{4.70}&{4.24}&{4.83}&{-26.68}&{4.76}&{-26.02}&{-12.41}&{6.30}&{-37.06}&{5.91}&{-36.36}&{5.99}\\
{1ENH}&{54}&{19}&{4.42}&{-0.67}&{4.75}&{-15.21}&{4.81}&{-10.8}&{-10.27}&{7.26}&{-38.85}&{7.68}&{-38.39}&{7.14}\\
{4PTI}&{58}&{32}&{6.18}&{-0.36}&{6.24}&{-8.03}&{6.06}&{-19.16}&{-6.95}&{7.00}&{-32.6}&{8.09}&{-35.65}&{8.62}\\
{2IGD}&{61}&{25}&{7.64}&{4.00}&{6.63}&{-18.21}&{6.53}&{-19.79}&{-10.28}&{9.4}&{-35.57}&{9.86}&{-36.49}&{8.69}\\
{1YPA}&{64}&{38}&{5.17}&{5.21}&{5.52}&{-26.90}&{5.39}&{-35.01}&{-17.1}&{8.37}&{-38.45}&{7.81}&{-40.14}&{8.32}\\
{1R69}&{69}&{30}&{4.44}&{3.59}&{4.76}&{-21.70}&{4.64}&{-22.37}&{-11.3}&{5.38}&{-39.89}&{7.16}&{-40.85}&{6.40}\\
{1CTF}&{74}&{42}&{4.72}&{-3.72}&{4.26}&{-32.55}&{4.08}&{-44.44}&{-18.06}&{7.19}&{-50.45}&{5.96}&{-51.50}&{5.94}\\
\hline
{3MX7}&{90}&{44}&{7.10}&{-0.08}&{7.21}&{-42.18}&{7.20}&{-50.85}&{-17.97}&{8.73}&{-56.55}&{10.05}&{-56.32}&{9.57}\\
{3NBM}&{108}&{56}&{5.89}&{-5.07}&{5.64}&{-35.75}&{5.37}&{-36.51}&{-23.09}&{8.27}&{-55.38}&{6.75}&{-53.66}&{7.34}\\
{3MQO}&{120}&{68}&{6.44}&{5.96}&{6.33}&{-51.44}&{6.38}&{-41.69}&{-15.47}&{9.31}&{-62.65}&{7.69}&{-62.25}&{8.13}\\
{3MRO}&{142}&{63}&{7.76}&{-10.97}&{7.93}&{-50.69}&{7.64}&{-68.41}&{-28.63}&{12.96}&{-90.56}&{11.89}&{-90.05}&{9.28}\\
{3PNX}&{160}&{84}&{7.90}&{-1.16}&{8.04}&{-73.90}&{7.60}&{-69.52}&{-26.79}&{10.81}&{-96.98}&{10.11}&{-102.55}&{10.12}\\
\hline
{3MSE}&{180}&{83}&{20.24}&{-14.41}&{16.05}&{-76.99}&{16.98}&{-77.73}&{-30.4}&{22.01}&{-91.02}&{19.12}&{-92.61}&{17.88}\\
{3MR7}&{189}&{88}&{10.43}&{-12.34}&{9.42}&{-84.28}&{9.36}&{-81.9}&{-26.99}&{10.56}&{-94.93}&{11.67}&{-93.65}&{10.84}\\
{3MQZ}&{215}&{115}&{11.21}&{-5.26}&{8.88}&{-98.75}&{9.04}&{-92.85}&{-15.51}&{11.53}&{-108.38}&{10.58}&{-104.29}&{10.7}\\
{3NO3}&{238}&{102}&{14.49}&{-14.51}&{11.22}&{-112.14}&{11.7}&{-100.79}&{-16.41}&{14.89}&{-119.9}&{13.2}&{-122.97}&{13.04}\\
{3NO6}&{248}&{112}&{13.2}&{-8.67}&{11.88}&{-120.23}&{12.06}&{-116.51}&{-44.07}&{13.96}&{-125.68}&{14.26}&{-133.95}&{13.09}\\
{3ON7}&{280}&{135}&{13.19}&{28.47}&{11.84}&{-105.63}&{11.77}&{-98.31}&{-8.59}&{13.95}&{-120.16}&{13.01}&{-116.88}&{16.58}\\

\hline
\end{tabular}	
\end{scriptsize}
\label{tabMinRMSD}
\end{table*}

\subsection{Result Analysis}
The MJ energy model actually implicitly bear the characteristic of hydrophobicity. The matrix values present some variations within amino acids of the same class (H or P). A partition algorithm such as 2-means clustering algorithm easily reveals the H-P partitioning within the MJ model. Given this knowledge, we study the effect of explicitly using hydrophobic property within our GA.

\subsubsection*{Effect of HP in MH model}
 Our macro-mutation operator biases the search towards a hydrophobic core by applying a series of diagonal moves and thus achieves improvements in terms of MJ energy values of the output conformations. We implemented three different versions of our genetic algorithm.
\begin{enumerate}
	\item {\bf MH:} This version is our final algorithm that we described in detail, and used in presenting our main results in {\tab}\ref{tabSOTAEnergy} and in comparing with the {\sota} results. To reiterate, this version uses the MJ energy model for search and energy reporting, and hydrophobicity knowledge in the macro-mutation operator that repeatedly applies diagonal moves towards forming a hydrophobic core.
	\item {\bf MJ:} This version of our GA uses the MJ energy model for search and energy reporting. This version has macro-mutation operator but not biased by hydrophobic properties of amino acids.
	\item {\bf HP:} This version of our GA uses the HP energy model for search. However, we report the energy values of the final conformations returned by the GA in MJ energy model. Note that this version has the hydrophobic core directed macro-mutation operator. This version will show whether HP model is sufficient even when the energy of a conformation is to be in the MJ model.
\end{enumerate}
From the Column RI in {\tab}\ref{tabMixedHP}, we see that MH guided GA improves the average interaction energy level over MJ model, ranging from $0.84$\% to $5.14$\% for all benchmark proteins. The improvements are not large in magnitudes but consistently better for all the proteins.
\begin{table}[!h]
\centering
\caption{\footnotesize The effect of using HP energy model within a macro-mutation operator.  The {bold-faced} values indicate the winners. The lower the energy value, the better the performance. The \emph{t-test} was performed with a confidence interval of 95{\pct}.}
\label{tabMixedHP}
\centering
\begin{scriptsize}
	\renewcommand{\arraystretch}{1.5}
	\setlength{\tabcolsep}{5pt}
	\begin{tabular}{cccrrrrrrr}
			\hline		
			\multicolumn{3}{c}{\bf Protein details}&\multicolumn{3}{c}{\bf Best of {\sf 50} runs}
			&\multicolumn{3}{c}{\bf Average[p-value] of {\sf 50} runs}&{\bf RI}\\
			\hline
			{\it Seq}&{\it Size}&{\it H}&{\it HP}&{\it MJ}&{\it MH }&{\it HP}&{\it MJ(\emph{r}) }&{\it MH(\emph{t})  }&{\it on MJ }\\
			\hline
			{4RXN}&{54}&{27}&{-12.41}&{\bf -37.71}&{-36.36}&{-3.54[2.4E-16]}&{-33.32[5.9E-56]}&\textbf{-33.60}[1.7E-75]&{0.84\%}\\
{1ENH}&{54}&{19}&{-10.27}&{-37.37}&{\bf -38.39}&{-7.29[3.8E-32]}&{-34.86[1.1E-66]}&\textbf{-35.67}[1.2E-70]&{2.32\%}\\
{4PTI}&{58}&{32}&{-6.95}&{-35.31}&{\bf -35.65}&{-2.81[1.5E-14]}&{-30.93[3.6E-55]}&\textbf{-31.01}[4.8E-67]&{0.26\%}\\
{2IGD}&{61}&{25}&{-10.28}&{\bf -36.97}&{-36.49}&{-6.75[2.7E-31]}&{-33.65[3.5E-66]}&\textbf{-33.75}[4.0E-70]&{0.30\%}\\
{1YPA}&{64}&{38}&{-17.1}&{-39.13}&\bf {-40.14}&{-9.90[2.3E-33]}&{-35.20[6.4E-65]}&\textbf{-36.33}[2.8E-73]&{3.21\%}\\
{1R69}&{69}&{30}&{-11.3}&{-39.77}&{\bf -40.85}&{-4.31[5.6E-19]}&{-35.43[4.9E-65]}&\textbf{-36.28}[2.5E-68]&{2.40\%}\\
{1CTF}&{74}&{42}&{-18.06}&{-50.09}&{\bf -51.5}&{-10.97[1.1E-32]}&{-44.98[1.4E-61]}&\textbf{-47.29}[6.8E-70]&{5.14\%}\\
\hline
{3MX7}&{90}&{44}&{-17.97}&{-55.57}&{\bf -56.32}&{-11.16[1.9E-31]}&{-48.46[5.5E-62]}&\textbf{-50.95}[2.6E-70]&{5.14\%}\\
{3NBM}&{108}&{56}&{-23.09}&{\bf -57.17}&{-53.66}&{-15.29[9.8E-36]}&{-48.47[9.5E-60]}&\textbf{-49.90}[2.6E-70]&{2.95\%}\\
{3MQO}&{120}&{68}&{-15.47}&{-60.22}&{\bf -62.25}&{-6.75[1.7E-18]}&{-53.00[4.8E-61]}&\textbf{-54.56}[2.4E-66]&{2.94\%}\\
{3MRO}&{142}&{63}&{-28.63}&{\bf -93.77}&{-90.05}&{-18.65[7.2E-31]}&{-79.32[2.1E-62]}&\textbf{-82.32}[1.6E-67]&{3.78\%}\\
{3PNX}&{160}&{84}&{-26.79}&{-99.87}&{\bf -102.55}&{-18.55[1.2E-34]}&{-85.64[6.0E-60]}&\textbf{-88.06}[1.3E-60]&{2.83\%}\\
\hline
{3MSE}&{180}&{83}&{-30.4}&{-91.02}&{-92.61}&{-13.17[5.0E-21]}&{-84.47[3.2E-70]}&{\textbf{-84.60}[3.6E-69}&{{0.20\pct}}\\
{3MR7}&{189}&{88}&{-26.99}&{-94.93}&{-93.65}&{-5.54[1.4E-06]}&{\textbf{-85.70}[4.1E-69]}&{-83.93[1.9E-36]}&{{\it non}}\\
{3MQZ}&{215}&{115}&{-15.51}&{-108.38}&{-104.29}&{6.86[8.7E-08]}&{\textbf{-96.58}[1.5E-68]}&{-95.22[6.7E-64]}&{{\it non}}\\
{3NO3}&{238}&{102}&{-16.41}&{-119.9}&{-122.97}&{-2.41[5.1E-02]}&{-108.68[1.1E-68]}&{\textbf{-108.70}[3.3E-65]}&{{0.12\pct}}\\
{3NO6}&{248}&{112}&{-44.07}&{-125.68}&{-133.95}&{-12.65[2.0E-11]}&{-116.31[1.8E-71]}&{\textbf{-117.11}[7.0E-67]}&{{0.70\pct}}\\
{3ON7}&{280}&{135}&{-8.59}&{-120.16}&{-116.88}&{9.38[7.0E-10]}&{\textbf{-104.57}[1.1E-56]}&{-96.64[2.4E-45]}&{{\it non}}\\
\hline
\end{tabular}	
\end{scriptsize}
\end{table}

\subsubsection*{Statistical significance}
We know that the lower \emph{p-values} are better. We performed the \emph{t-test} with a confidence interval of 95{\pct} (i.e., significance level is 5{\pct}) and the results are presented in {\tab}\ref{tabMixedHP}. For MJ and MH models, the p-values of all proteins are less than the significance level. However, for HP model, the p-value for 3MSE is below the significance level and for other five sequences those are equal to the significance level. Therefore, the experimental results are statistically significant.

\subsubsection*{Search progress}
\begin{figure}[!h]
\centering
\begin{tabular}{cc}
	{\includegraphics[width=8cm]{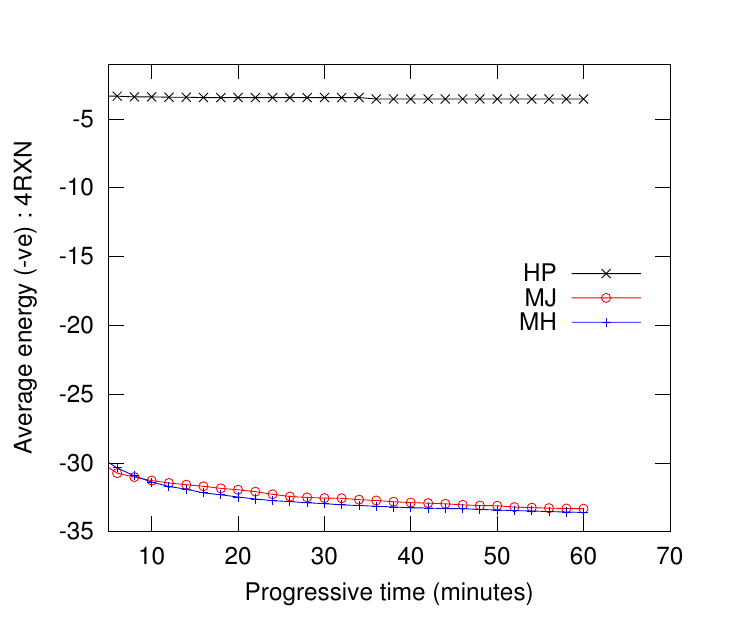}}&
	{\includegraphics[width=8cm]{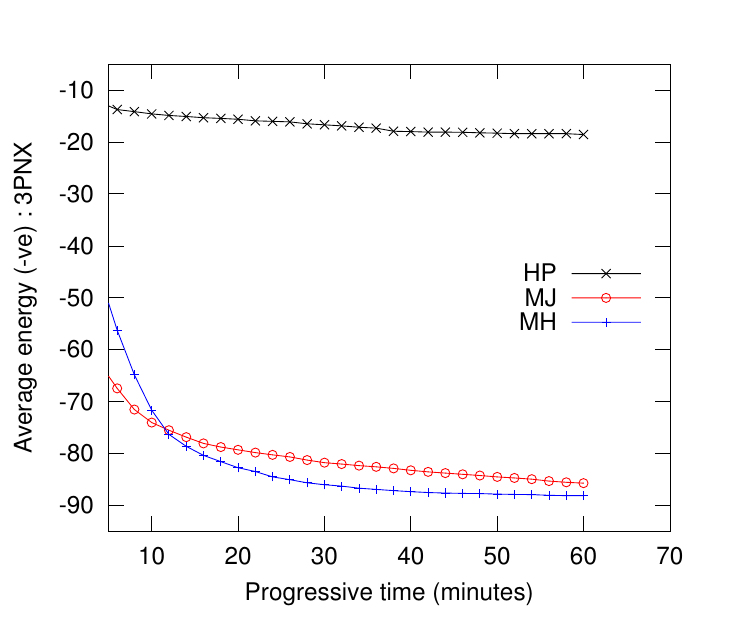}}
\end{tabular}
\vspace{-2ex}
\caption{\footnotesize  The search progress over a time-span of $60$ minutes for proteins \emph{4RXN} and \emph{3PNX} of sequence length $54$ and $160$ amino acids respectively.}
\label{figSearchProgress}
\end{figure}

To demonstrate the search progress, we periodically find the best energy values obtained so far in each run. For a given period, we then calculate the average energy values obtained for that period over 50 runs. We used a 2 minute time interval. {\fig}\ref{figSearchProgress} presents the average energy values obtained at each time interval for two different proteins: {\ba} and {\ce} are the smallest and largest amongst the 12 benchmark proteins respectively. From both of the charts, we see that the final version of our algorithm MH performs better than the other two versions.

\section{Discussion}
\label{secDiscussion}
By encoding the conformation with angular coordinates ($\phi$ and $\psi$), our GA might easily be applied in high-resolution PSP. While the minimizing energy function is highly complex (such as molecular dynamics), a simple guidance heuristic---such as hydrophobic property or exposed surface area---could be used to guide the macro-mutation operator. Within GA framework, the macro-mutation operator could be applied optimizing the segments of secondary structures ($\alpha-$helix and $\beta-$sheet).

Our approach can easily divide the whole optimization process into two stages guided by two energy models with different complexities. The macro-mutation operator can be guided by simpler energy models such as distance from hydrophobic core, exposed surface area, hydrophobicity of amino acids, hydropathy index of the amino acids, and so on. Conversely, the main objective function can be more realistic such as molecular dynamics based energy models. This two-stage optimization will reduce the overall computational complexities. As a result, our framework has a good chance to succeed in more realistic models even for large sized proteins.

\section{Conclusion}
\label{secConclusion}

Our guided macro-mutation in a graded energy based genetic algorithm, `MH\_GeneticAlgorithm', is found to be an effective sampling algorithm for the convoluted protein structure space. The strategical switching in between the Miyazawa-Jernigan (MJ) energy and the Hydrophobic-Polar (HP) energy made the proposed algorithm perform better compared to the other state-of-the-art approaches. This is because, while the fine graded MJ energy interaction computation become computationally prohibit, the low resolution HP energy model can effectively sample the search-space towards certain promising directions. In addition, the GA framework was enhanced and made powerful, since it uses not only crossover but also three effective move operators. Further to diversify the population to keep sampling or, exploring the search space effectively, a hydrophobic core-directed macro-mutation operator, twin removal as well as a random-walk algorithm to recover from the stagnation have been applied. To compare the performance of our GA, we have extensively compared with the existing state-of-the-approaches using the benchmark problem available and found our approach to be consistently better and often found significantly better and t-test result in terms of p-values have been provided. For the lattice configuration to be followed, we used 3D face-centered-cube (FCC) lattice model because prediction in the FCC lattice model can yield the densest protein core and the FCC lattice model can provide the maximum degree of freedom as well as the closest resemblance to the real or, high resolution folding within the lattice constraint. This enables the predicted structure to be aligned and hence, migrated to a real protein (prediction) model efficiently for future extensions.

\section*{Acknowledgment}
Mahmood Rashid and Abdul Sattar would like to express their great appreciation to National ICT Australia. Sumaiya Iqbal and Md Tamjidul Hoque acknowledge the Louisiana Board of Regents through the Board of Regents Support Fund, LEQSF(2013-16)-RD-A-19.

\section*{References}
\begin{spacing}{1.0}
	\begin{small}
		\bibliography{psp_cbac}
	\end{small}
\end{spacing}

\end{document}